\documentclass[lettersize,journal]{IEEEtran}
\usepackage{amsmath,amsfonts}
\usepackage{algorithmic}
\usepackage{algorithm}
\usepackage{array}
\usepackage{textcomp}
\usepackage{stfloats}
\usepackage{url}
\usepackage{verbatim}
\usepackage{graphicx}
\usepackage{cite}
\usepackage{booktabs}
\usepackage{multirow}
\usepackage{makecell}
\usepackage{bm}
\usepackage[]{hyperref}
\hypersetup{hidelinks}
\usepackage{subcaption}
\usepackage{color}
\usepackage{arydshln}

\begin{document}
\title{Personalized Cross-Modal Emotional Correlation Learning for Speech-Preserving Facial Expression Manipulation}

\author{Tianshui Chen, Yujie Zhu, Jianman Lin, Zhijing Yang, Chunmei Qing, Feng Gao, and Liang Lin,~\IEEEmembership{Fellow,~IEEE} % <-this % stops a space
\thanks{Tianshui Chen, Yujie Zhu, and Zhijing Yang are with Guangdong University of Technology (Emails: tianshuichen@gmail.com, 2112303087@mail2.gdut.edu.cn, yzhj@gdut.edu.cn). Jianman Lin and Chunmei Qing are with South China University of Technology (Emails: linjianmancjx@gmail.com, qchm@scut.edu.cn). Feng Gao is with Peking University (Email: gaof@pku.edu.cn). Liang Lin is with Sun Yat-Sen University (Email: linliang@ieee.org). Feng Gao is the corresponding author.
This work was supported by the Natural Science Foundation of Guangdong Province(Grant 2025A1515010454)}% <-this % stops a space
}

% \thanks{Manuscript received April 19, 2021; revised August 16, 2021.}

% The paper headers
% \markboth{Journal of \LaTeX\ Class Files,~Vol.~14, No.~8, August~2021}%
% {Shell \MakeLowercase{\textit{et al.}}: A Sample Article Using IEEEtran.cls for IEEE Journals}

%\IEEEpubid{0000--0000/00\$00.00~\copyright~2021 IEEE}
% Remember, if you use this you must call \IEEEpubidadjcol in the second
% column for its text to clear the IEEEpubid mark.

\maketitle

\begin{abstract}
\textcolor{black}{Speech-preserving facial expression manipulation (SPFEM) aims to enhance human expressiveness without altering mouth movements tied to the original speech. A primary challenge in this domain is the scarcity of paired data—aligned frames of the same individual with identical speech but differing expressions—which impedes direct supervision for emotional manipulation.
While current Visual-Language Models (VLMs) can extract aligned visual and semantic features, making them a promising source of supervision, their direct application is limited. To this end, we propose a Personalized Cross-Modal Emotional Correlation Learning (PCMECL) algorithm that refines VLM-based supervision through two major improvements.
First, standard VLMs rely on a single, generic prompt for each emotion, failing to capture expressive variations among individuals. PCMECL addresses this by conditioning on individual visual information to learn personalized prompts, thereby establishing more fine-grained visual-semantic correlations. Second, even with personalization, inherent discrepancies persist between the visual and semantic feature distributions. To bridge this modality gap, PCMECL employs feature differencing to correlate the modalities, providing more precisely aligned supervision by matching the change in visual features to the change in semantic features.
As a plug-and-play module, PCMECL can be seamlessly integrated into existing SPFEM models. Extensive experiments across various datasets demonstrate the superior efficacy of our algorithm.}

\end{abstract}


\section{Introduction}
\IEEEPARstart{S}{peech-preserving} facial expression manipulation (SPFEM) is a crucial technology that modifies facial expressions without affecting the mouth movements synchronized with speech. This innovation significantly boosts human expressiveness and offers substantial advantages for applications like virtual avatars and film and television production. Traditionally, achieving accurate emotional portrayal in film requires extensive visual effects and numerous retakes, which are both costly and time-consuming. Conversely, an advanced SPFEM system can efficiently alter expressions in post-production, achieving high-quality results more easily and effectively. These systems are eagerly awaited for their ability to enhance creative processes and deepen the emotional resonance of visual media.

\begin{figure}[t]
\centering
\includegraphics[width=0.9\columnwidth]{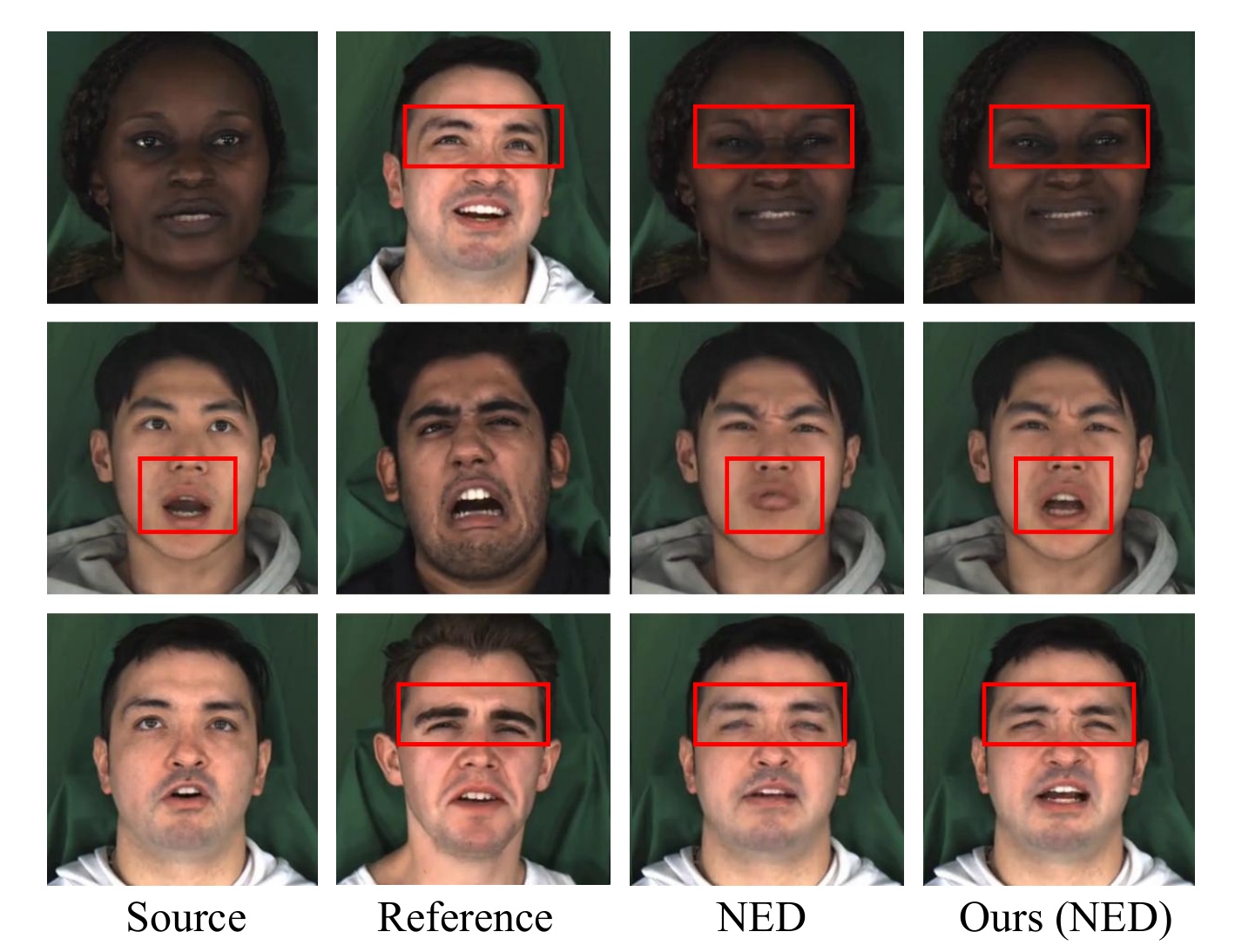} 
\caption{
\textcolor{black}{
Motivating examples showing baseline (NED) failures and our improvements. From left to right: source identity image, reference emotion image, the baseline's result, and our result. Red boxes highlight key areas for comparison.
}
}
\label{fig: motivation_figure}
\end{figure}

\textcolor{black}{
A fundamental challenge in advancing Speech-Preserving Facial Expression Manipulation (SPFEM) is the scarcity of paired supervision. The logistical difficulty of collecting large-scale datasets where an individual utters identical speech with varied emotions makes direct training intractable. Early attempts \cite{doukas2021head2head++, tripathy2020icface, li20223d3m, chai2021expression} sought to bypass this limitation by leveraging face reenactment techniques, using reference images for frame-by-frame guidance. However, this paradigm is often plagued by attribute leakage, where unintended features from the reference image compromise the fidelity of the target expression. Subsequent methods \cite{papantoniou2022neural} employed cycle consistency to establish correspondence between unpaired images, yet the reliance on global constraints struggles to model the subtle, localized facial changes inherent to emotional expression. Consequently, existing approaches face a critical trade-off: they either fail to accurately convey the intended emotion or struggle to preserve crucial, speech-synchronized mouth movements, as shown in Fig. \ref{fig: motivation_figure}. This highlights a key open challenge: the need to construct effective, fine-grained supervision signals in the absence of paired data.}

\begin{figure}
    \centering
    \begin{subfigure}{0.23\textwidth} 
        \includegraphics[width=\textwidth]{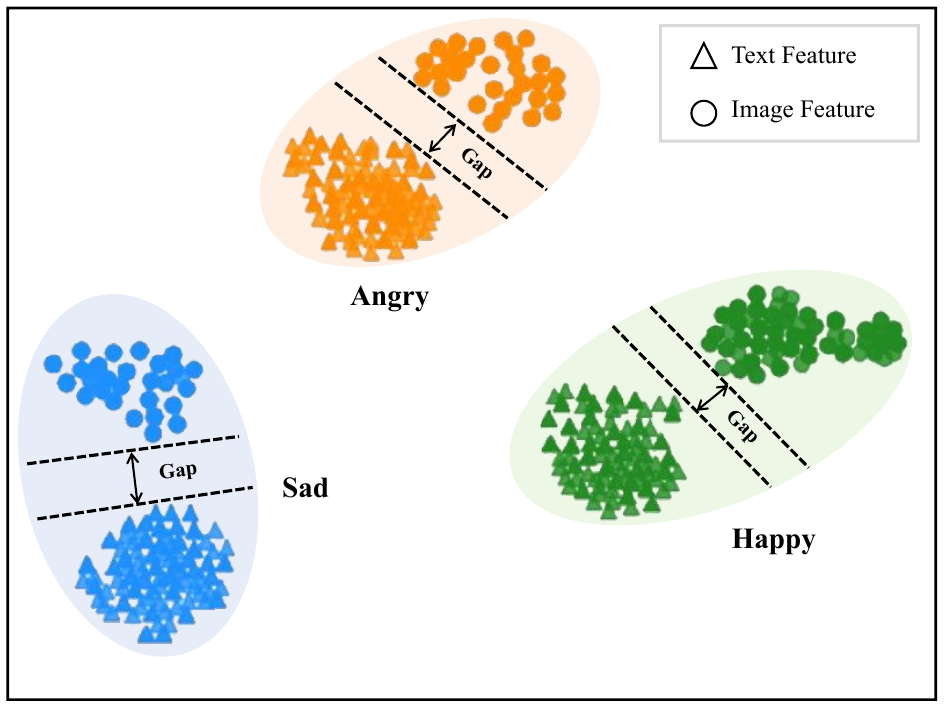} 
        \caption{} 
        \label{fig:CMECC-1}
    \end{subfigure}
    \hfill
    % \hspace{1pt}
    \begin{subfigure}{0.23\textwidth} 
        \includegraphics[width=\textwidth]{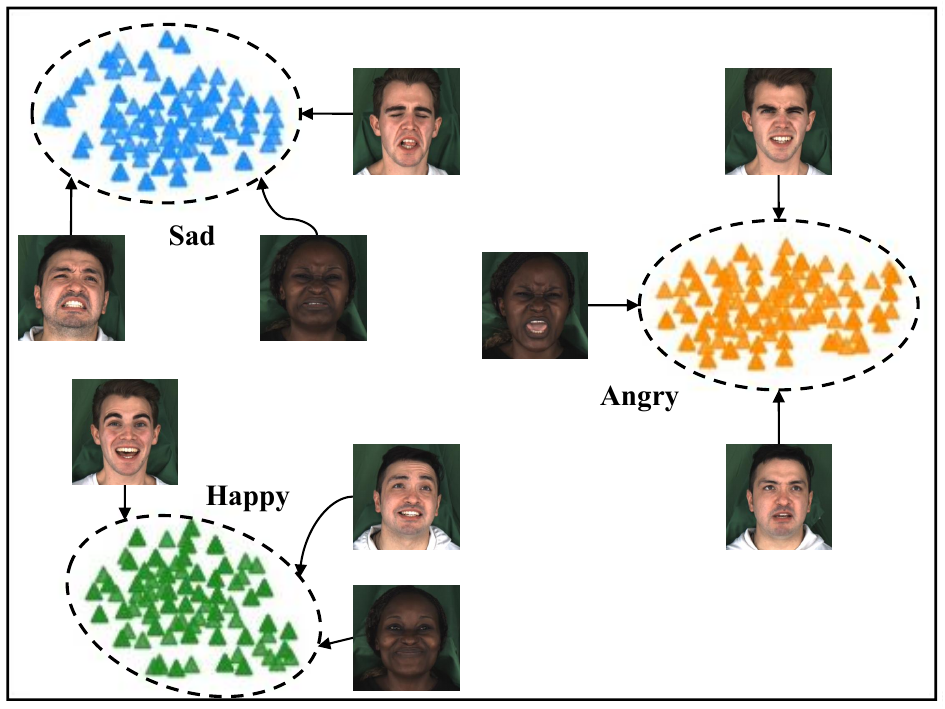} 
        \caption{} 
        \label{fig:CMECC-2}
    \end{subfigure}
    \vspace{-5pt}
    % \caption{The distribution of visual features and text features. (a) The disparity in modality between visual and text features results in inherent gap in their feature distribution. (b) The current VLMs uses static prompts to represent emotional features.}
    \caption{
    \textcolor{black}{
    \textbf{Challenges in Applying Pre-trained VLMs, visualized via t-SNE of CLIP features from MEAD.} 
    \textbf{(a)} The \textbf{inherent modality gap} between image (circles) and text (triangles) feature distributions, which remain structurally separated. 
    \textbf{(b)} The \textbf{limitation of fixed prompts}: current VLMs use a "one-size-fits-all" approach, mapping diverse visual expressions of an emotion to a single cluster defined by a fixed text prompt, thereby failing to capture individual styles.
    }
    }
    \label{fig: CMECC}
\end{figure}

\textcolor{black}{Intuitively, the rise of Visual-Language Models (VLMs) \cite{radford2021learning, alayrac2022flamingo} presents a promising avenue. By aligning visual features with rich semantic representations, they offer a potential source of supervision for generating images that match a target emotion. However, directly leveraging off-the-shelf VLMs for SPFEM is hindered by two fundamental challenges.
First, standard VLMs employ generic, one-size-fits-all prompts for emotions. As depicted in Fig. \ref{fig: CMECC} (b), this approach fails to capture the nuanced and unique ways different individuals express the same emotion, overlooking critical expressive variations. Second, and more fundamentally, even if personalization were addressed, a significant modality gap persists between the visual and text feature spaces. Due to their inherent structural differences, their distributions remain misaligned even after contrastive pre-training, as illustrated in Fig. \ref{fig: CMECC} (a). This misalignment leads to imprecise and unreliable supervision signals.}

\textcolor{black}{In this work, we propose the Personalized Cross-Modal Emotional Correlation Learning (PCMECL) algorithm, a framework designed to generate fine-grained and robust supervision for SPFEM methods by refining cutting-edge VLMs \cite{radford2021learning}. Our approach begins by addressing the lack of personalization through a Personalized Emotional Prompt Learning (PEPL) module.
Specifically, PEPL employs a textual inversion-like process to embed an individual's unique appearance into the VLM's semantic space. This technique learns a new, personalized token that represents the subject's identity, which is then combined with standard emotion prompts (e.g., "happy," "sad"). By conditioning on this identity-aware prompt, the PEPL module effectively captures the expressive variations unique to each subject and yields highly specific, fine-grained correlations between visual expressions and their semantic representations.
With these personalized features, we then tackle the underlying modality gap using Visual-Text Emotional Differencing Correlation (VTEDC) regularization. Instead of directly matching disparate feature distributions, VTEDC provides supervision by enforcing consistency between their differences. During SPFEM model training, we correlate the change between the source and target visual representations with the corresponding change between their semantic representations. This differencing technique provides a precise and stable supervisory signal, ensuring more accurate and expressive emotional manipulation.}

\textcolor{black}{The contribution can be summarized into four folds:} 
\begin{itemize}
    \item \textcolor{black}{We pioneer the use of VLM-based supervision for the SPFEM task, a contribution that also involves being the first to systematically identify and address the dual challenges of the modality gap and lack of personalization.}
    % \item \textcolor{black}{We are the first to identify and address the dual challenges of the modality gap and lack of personalization in VLM-based supervision for SPFEM.}
    \item \textcolor{black}{We introduce the PEPL module, which solves the personalization problem by utilizing individual visual information to learn personalized emotional prompts.}
    \item \textcolor{black}{We propose the VTEDC module, which pioneers a new difference alignment paradigm to bypass the modality gap by aligning feature differences rather than absolute features.}
    \item \textcolor{black}{We demonstrate through extensive experiments that our plug-and-play PCMECL framework significantly enhances the performance of SPFEM models.}
\end{itemize}

\section{Related Works}
\subsection{Speech-preserving facial expression manipulation}
Traditional face reenactment \cite{zhang2024emotalker, eskimez2021speech, kim2019neural, ren2021pirenderer, wang2022latent, xu2023progressive, zheng2023face, xu2025exploiting, gong2025monocular} involve an actor replicating both the speech and facial expressions as depicted in a reference video. The main methods for face reenactment involve the parameterization of facial images \cite{doukas2021head2head++, tu20203d, chen2020self, gan2023efficient, siarohin2019first, chen2024learning, chen2025contrastive}, such as 3DMM parameters and action units, and then adjusting these parameters to achieve the desired facial configurations. For instance, ICface \cite{tripathy2020icface} manipulates the pose and expression of a given face image through interpretable control signals, which include head pose angles and action unit values. GANimation \cite{pumarola2018ganimation} utilizes adversarial learning, guided by action unit (AU) annotations \cite{ekman1978facial}, to represent facial movements in a continuous space. This method allows for the fine-tuning of each AU's activation intensity and the blending of multiple AUs. Concurrently, there are some works \cite{yin2022styleheat, choi2018stargan, hang2023language, karaouglu2021self, tewari2020stylerig, sun2023continuously, xu2024self} employ StyleGAN \cite{goodfellow2014generative, karras2019style} to convert facial images into latent space and then modify latent code to achieve face reenactment. For example, StyleHEAT \cite{yin2022styleheat} extends StyleGAN's latent code to enable motion and expression animation by incorporating video-based and audio-based motion generation modules, along with a calibration network to correct transformation distortions. While current face reenactment technology has advanced significantly, it still falls short of meeting the requirements for the SPFEM task, which demands the synchronization of manipulated facial expressions with the oral movements of the original audio. 

SPFEM focuses on transforming a source video to express a specific emotion while retaining the facial movements aligned with the speech content. For this \textcolor{black}{purpose}, previous studies have leveraged facial reenactment techniques. For instance, ICface \cite{tripathy2020icface} enables control over head poses and facial expressions by adjusting head pose angles and action units. However, these methods, while adept at replicating expressions and mouth shapes, often fail to retain the essence of speech. Wav2Lip-Emotion \cite{magnusson2021invertable} enhances traditional lip synchronization architectures by \textcolor{black}{incorporating} advanced emotional modulation capabilities. This approach achieves facial emotion manipulation through a combination of L1 reconstruction loss and pre-trained emotion objectives, ensuring accurate synchronization of lip movements while integrating expressive emotional cues into the generated outputs. \textcolor{black}{Nevertheless, these methods face} significant challenges in maintaining the consistency of facial identity in test images, often resulting in noticeable discrepancies between the generated and original appearances. These limitations hinder their applicability in scenarios where preserving facial identity and achieving high-fidelity visual representation are critical. NED \cite{papantoniou2022neural} deals with this shortcoming by merging 3DMM parameters for source identification with those for target emotion expression. 

\textcolor{black}{
In addition to these methods, recent frameworks like SSERD~\cite{xu2024self} have made significant strides by creating synthetic paired data for pseudo-supervised training. Similarly, learning spatial-temporal coherent correlations has recently been proposed to better preserve localized facial dynamics during expression manipulation \cite{chen2026learning}. The success of these approaches underscores a critical insight: access to fine-grained, paired-like supervision is key to achieving high-quality results. This motivates our work to explore a novel paradigm for generating such supervision directly from real data using VLMs, rather than relying on synthetic data generation.
}

\textcolor{black}{
Despite these advancements, most methods still lack the mechanisms for precise and personalized emotional supervision, a key challenge that remains open in the SPFEM domain.
}

\subsection{Visual-language model}
Visual-Language Models (VLMs) \cite{radford2021learning, alayrac2022flamingo, li2022blip, kwon2022masked} represent a powerful integration of techniques from both computer vision and natural language processing, enabling the processing of both visual content and associated textual information. These models are typically trained on vast datasets, which allows them to develop a robust capacity for semantic understanding across multiple domains. Their advanced semantic capabilities position VLMs to serve as foundational models for a variety of tasks \cite{zhou2022conditional, wu2023high, zhao2023prompting, xenos2024vllms}, or to be fine-tuned for specific downstream tasks \cite{patashnik2021styleclip, yang2023zero, li2023cliper, etesam2024contextual}, enhancing performance through additional supervision. 

For example, research like \cite{xu2023high} employs the CLIP model \cite{radford2021learning} to leverage contrastive learning, establishing meaningful connections between images and text in a shared embedding space. This approach has proven effective for generating emotional talking head models that convey lifelike expressions and behaviors. Similarly, work such as \cite{gal2022stylegan} demonstrates how CLIP can be used to adapt pre-trained image generators across domains. By leveraging CLIP's semantic understanding, the method enables text-driven image editing without requiring extensive training data, extending the applicability of generative models to new domains. In \cite{chefer2022image}, a novel method is introduced for transferring high-level semantic attributes from a target image to a source image. This method integrates the StyleGAN generator with CLIP's semantic encoder, ensuring both identity preservation and high-level feature transfer. It surpasses existing techniques in style transfer, domain adaptation, and text-based semantic editing, underscoring CLIP's versatility in guiding complex image transformations. These approach highlights CLIP's potential in bridging semantic understanding and generative modeling for flexible domain adaptation.

% Current VLMs can extract aligned visual and emotional features through their visual and semantic encoders, allowing these features to effectively guide the generation of visual images that are more aligned with the target emotion. However, their effectiveness is limited by cross-modal gaps in visual-textual feature distribution and individual differences in emotional expression patterns, as observed in the SPFEM task. PCMECL mitigates these modality discrepancies by aligning visual and textual features in a more consistent manner through the VTEDC module, and focuses on extracting fine-grained, personalized information from the data through the PEPL module, effectively applying VLM to the training of the SPFEM model.
\textcolor{black}{
These works highlight the immense potential of VLMs for semantic image manipulation. However, directly applying them to fine-grained supervision tasks like SPFEM introduces specific challenges, such as the inherent modality gap and the need for personalization, which have not been systematically addressed.
}

\begin{figure*}[htp]
  \centering
  \vspace{-25pt}
  \includegraphics[width=\textwidth]{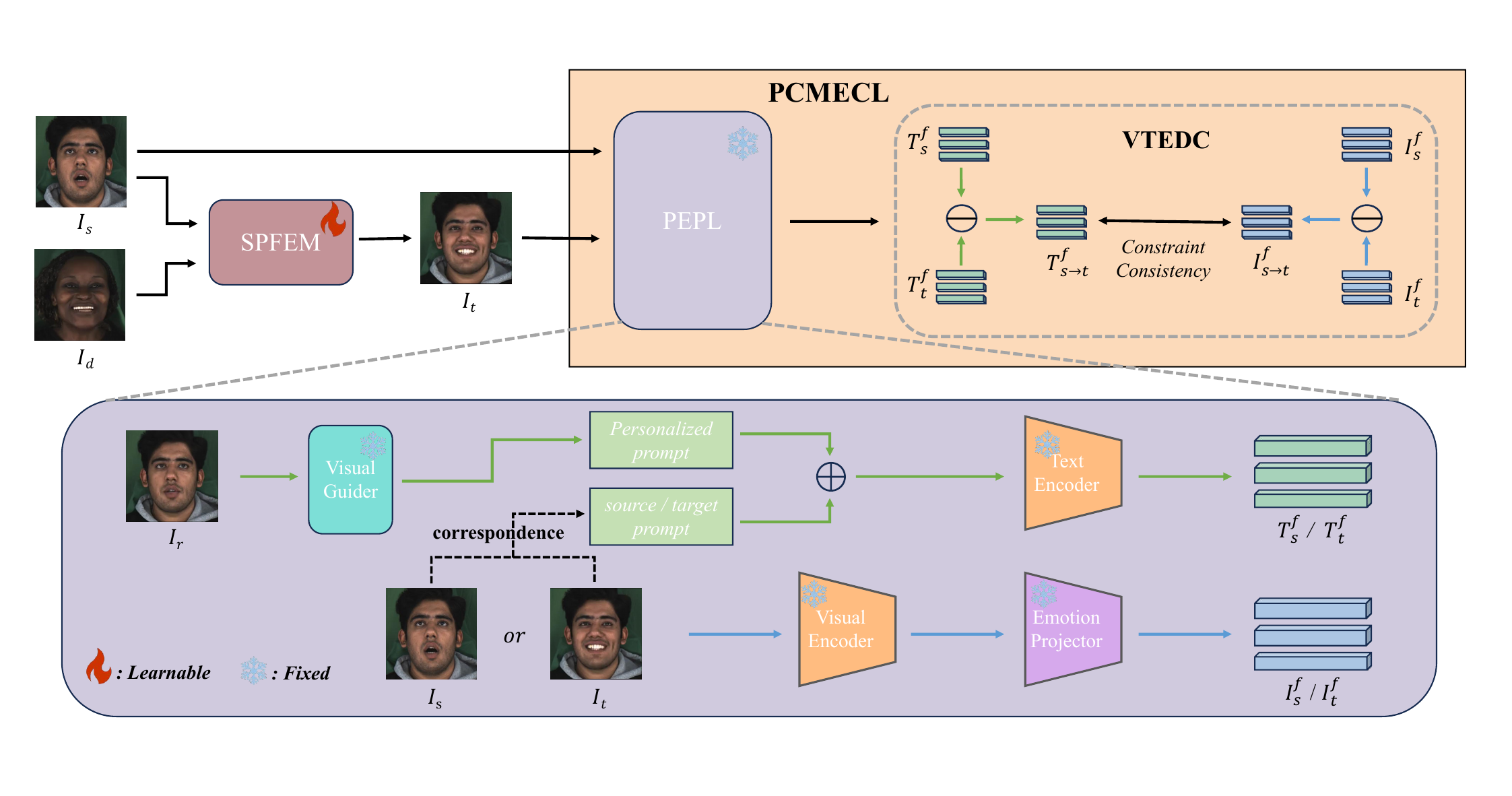} 
  \vspace{-25pt}
  % \caption{ The overall architecture of incorporating the proposed PCMECL algorithm into the SPFEM model to supervise the generation of facial image. The purple box indicates the pre-trained PEPL module, designed for harvesting emotional visual-text feature pairs. The dotted line box indicates the process of VTEDC, where the blue and green lines respectively demonstrate the information flow for visual and text feature extraction, designed to bridge the inherent disparities between visual and text features.}
  % \caption{
  % \textcolor{black}{
  % \textbf{The overview of PCMECL supervisory framework.} During the training of an SPFEM model, PCMECL takes the source inputs \(I_s\) and the target outputs \(I_t\) as its inputs. The pre-trained and frozen PEPL modules extract personalized visual (\(I_f\)) and textual (\(T_f\)) embeddings for both the source and target. The core VTEDC mechanism then aligns the difference vectors (\(I^f_{s \to t}\) and \(T^f_{s \to t}\)) by enforcing directional consistency. This consistency constraint serves as a supervisory signal that is backpropagated to optimize the SPFEM model. The reference image \(I_r\) provides identity information for personalization within the PEPL modules.
  % }
  % }
\caption{
\textcolor{black}{
\textbf{Overview of the PCMECL supervisory framework.} 
During SPFEM training, the frozen PCMECL module computes a supervisory loss from the source (\(I_s\)) and target (\(I_t\)) images. 
Its PEPL module process each image via two parallel branches: 
a \textbf{visual branch} extracts an emotion-centric visual embedding (\(I^f\)), while a \textbf{textual branch} uses a neutral reference image \(I_r\) to create a personalized text embedding (\(T^f\)). 
The VTEDC mechanism then aligns the difference vectors between the source and target embeddings (\(I^f_{s \to t}\) and \(T^f_{s \to t}\)) to provide a supervisory signal for optimizing the SPFEM model.
}
}
  % \vspace{-15pt}
  \label{fig: Framework}
\end{figure*}

\begin{figure}[t]
\centering
\includegraphics[width=0.9\columnwidth]{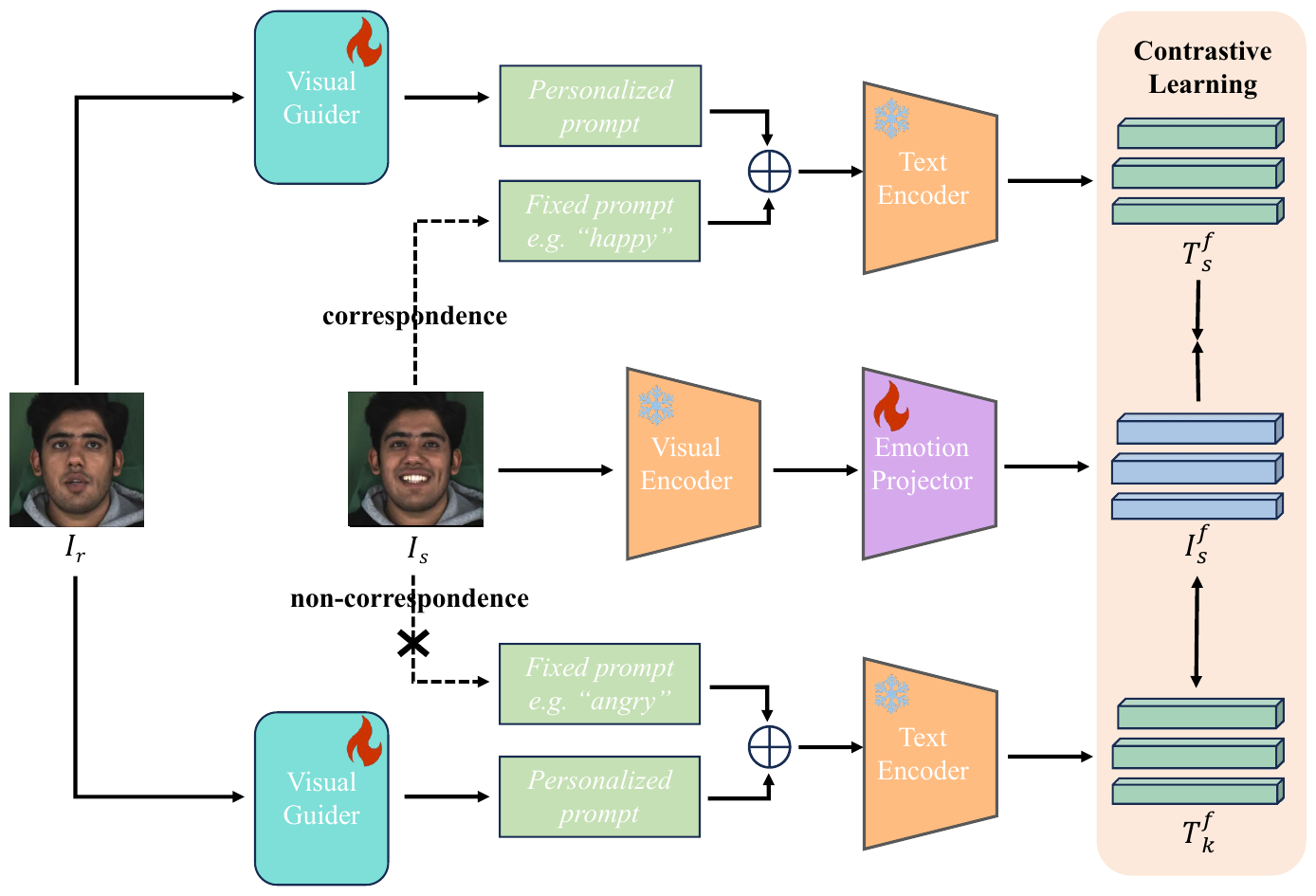} 
% \caption{An illustration of PEPL module. It depicts the process of extracting emotional visual-text feature pairs.}
\caption{
\textcolor{black}{
\textbf{Overview of the PEPL module's contrastive pre-training.} 
The learnable Emotion Projector extracts an emotion-centric visual embedding \textcolor{black}{\(I_s^f\)} from an emotional image \(I_s\). Concurrently, the learnable Visual Guider extracts a personalized embedding from a reference image \(I_r\), which is then concatenated with a corresponding text prompt to form the positive pair \textcolor{black}{\(T_s^f\)}, and with a non-corresponding prompt for the negative pair \textcolor{black}{\(T_k^f\)}. The training objective is to align the positive pair \textcolor{black}{\((I_s^f, T_s^f)\)} while separating the negative pair \textcolor{black}{\((I_s^f, T_k^f)\)}.
}
}
\vspace{-8pt}
\label{fig: EITRL}
\end{figure}

\section{Preliminaries}
\label{PCMECL:preliminaries}

\textcolor{black}{
Our proposed method, PCMECL, serves as a plug-and-play supervisory framework for existing Speech-Preserving Facial Expression Manipulation (SPFEM) models. In our experiments, we demonstrate its effectiveness on two representative baselines with distinct underlying mechanisms: NED~\cite{papantoniou2022neural} and ICface~\cite{tripathy2020icface}. To provide the necessary context for our work, we briefly introduce their core concepts to provide context for our contributions.
}

\textcolor{black}{
\paragraph{NED}
NED is a two-stage model that leverages a 3D Morphable Model (3DMM) for disentanglement. Its first stage analyzes the input video to extract 3DMM parameters representing identity and expression. The second stage, an Emotion Manipulator inspired by StarGANv2~\cite{choi2020v2}, translates the expression parameters to a target emotion before the final rendering. The training is guided by a composite loss function, \(L_{\text{NED}}\), which includes adversarial, style reconstruction, cycle consistency, and speech-preserving terms. This parameter-based manipulation offers explicit control but is highly dependent on the accuracy of the initial 3DMM parameter estimation.
}

\textcolor{black}{
\paragraph{ICface}
In contrast, ICface is a single-stage, GAN-based model for face reenactment, controlled via interpretable Action Units (AUs). It consists of two main generators: one to create a neutral face representation and another to perform the reenactment based on target AUs. Its composite loss function, \(L_{\text{ICface}}\), is designed around this GAN framework and includes adversarial, cycle consistency, AU control, and identity losses. The expressiveness of this approach is therefore tied to the quality of the driving AUs, which can be challenging to decouple from other facial attributes.
}

\textcolor{black}{
\paragraph{SSERD}
SSERD is a recent framework that focuses on self-supervised emotion representation disentanglement. A key innovation of SSERD is its paired data construction module, which leverages a pretrained StyleGAN to generate synthetic paired data. This allows for a form of detailed, pseudo-supervised training. The model is trained with a hybrid strategy, combining this synthetic paired data with real unpaired data to enhance realism.
}

\textcolor{black}{
\paragraph{Core Limitation of Baselines}
Despite their different architectures \textcolor{black}{and supervisory sources (e.g., the self-supervision in NED and ICface, and the pseudo-supervision in SSERD)}, all three baselines are fundamentally constrained by the lack of large-scale, \textbf{real} paired data for emotional supervision. Consequently, they rely on indirect, self-supervised signals (e.g., cycle-consistency). As we illustrate in our motivating examples (Fig.~\ref{fig: motivation_figure}), this weak form of supervision can only optimize for geometric plausibility, not semantic accuracy, leading to artifacts and inaccurate expressions. Our PCMECL framework is specifically designed to address this supervisory gap. For a detailed breakdown of the complete loss functions for both NED and ICface, please refer to Section I of the Supplementary Material.
}

\section{Method}
% The proposed Personalized Cross-Modal Emotional Correlation Learning (PCMECL) algorithm supervises the training of the SPFEM model by constructing and utilizing correlations between differences in visual and emotional representations, ensuring more accurate and expressive emotional manipulation. Specifically, It first pretrains a Personalized Emotional Prompt Learning (PEPL) module based on a visual-language model, enabling personalized text embedding and emotion-centric visual embedding generation. Then, it designs a Visual-Text Emotional Differencing Correlation (VTEDC) regularization strategy, which constrains the difference between the visual embedding pairs of the SPFEM source input and target output to be consistent with the difference between the text embedding pairs of the source and target emotions, providing fine-grained emotional supervision for the SPFEM model.  An overall pipeline for incorporating PCMECL into the SPFEM model is illustrated in Fig. \ref{fig: Framework}. \textcolor{black}{Our framework is built upon a pre-trained Vision-Language Model (VLM), for which we use CLIP with a ViT-B/32 backbone throughout our work.}

\subsection{Personalized Emotional Prompt Learning}
\label{PCMECL:pepl}
% PEPL enhances the current VLM by introducing a learnable Visual Guider and an Emotion Projector, which generate personalized text embeddings and emotion-centric visual embeddings, respectively. It then leverages cross-modal emotional correlation learning to align these embeddings by exploring the intrinsic relationships between the modalities within the VLM, as illustrated in Fig. \ref{fig: EITRL}.
\textcolor{black}{
To address the dual challenges of providing emotional supervision and ensuring it is personalized, which represents a key limitation of existing approaches, we propose the Personalized Emotional Prompt Learning (PEPL) module. PEPL is designed to overcome the "one-size-fits-all" nature of standard VLM supervision by generating identity-aware emotional prompts. It achieves this by introducing a learnable Visual Guider and an Emotion Projector by exploring the intrinsic relationships between the modalities within the VLM, as illustrated in Fig. \ref{fig: EITRL}.
}

\textcolor{black}{
The Visual Guider is responsible for capturing individual appearance information, which refers to the stable, identity-defining characteristics of a person extracted from a neutral reference image (\(I_r\)). This includes features like bone structure, inherent skin texture or subtle facial lines that define how a specific person uniquely expresses emotions. By using a neutral image, we enable effective feature disentanglement, providing a clean and unbiased identity baseline, free from the influence of any emotional muscle activation. This prevents the identity representation from being contaminated by features from an emotional reference, which would otherwise create conflicts and artifacts when generating a different target emotion. The Emotion Projector is designed to extract fine-grained visual features corresponding to a specific emotion from the source image \(I_s\).
}

% % Formally, given a facial image \(I_s\) and its corresponding and non-corresponding emotion text prompts \(T_s\) and \(T_k\), as well as a neutral facial image $I_r$.
% \textcolor{black}{Formally, for the PEPL pre-training stage, we utilize an emotional image \(I_s\), its corresponding text prompt \(T_s\), a non-corresponding prompt \(T_k\), and a neutral facial image \(I_r\) of the same identity as.}
% \textcolor{black}{The choice to use a neutral image for \(I_r\) is a crucial design decision to enable effective feature disentanglement. It provides a clean and unbiased identity baseline, free from the influence of any emotional muscle activation. This prevents the identity representation from being contaminated by features from an emotional reference (e.g., a smile), which would otherwise create conflicts and artifacts when generating a different target emotion.}
Our goal is to harness the inherent correlation between the visual and semantic modalities within a VLM to learn emotion-centric visual embeddings and personalized text embeddings. Recognizing that each individual expresses the same emotion uniquely, and acknowledging the difficulty of using existing pre-trained identity feature extractors to capture identity information and embed it into the latent space of the VLM due to domain gaps, 
% we propose employing textual inversion to generate personalized emotional prompts for each individual. 
\textcolor{black}{we propose a dynamic and conditional adaptation of textual inversion to generate personalized emotional prompts. While standard Textual Inversion learns a single, fixed pseudo-word for a visual concept, our approach trains the Visual Guider module to generate a unique, identity-aware embedding for each individual. This dynamically generated embedding functions as a conditional pseudo-word that represents the person's unique expressive style. Crafting this personalized emotional target signal is the core innovation of our PEPL module, which is then used to guide the SPFEM model.}

\textcolor{black}{Formally, for the PEPL pre-training stage, we utilize an emotional image \(I_s\), its corresponding text prompt \(T_s\), a non-corresponding prompt \(T_k\), and a neutral facial image \(I_r\) of the same identity as. Crucially, the PEPL module is trained via a cross-modal contrastive learning objective, not direct supervision using emotion labels. This objective leverages carefully constructed positive and negative pairs to learn discriminative features. The process to generate these personalized embeddings is formalized as follows.}

Specifically, we first tokenize the individual's appearance information using a Visual Guider and then concatenate it with the tokenized results of the corresponding and non-corresponding emotion text prompts:
\begin{align}
\begin{split}
T_{s}^{p} &= \text{concat}(E(I_r), F_t(T_s)) \\
T_{k}^{p} &= \text{concat}(E(I_r), F_t(T_k))
\end{split}
\end{align}

Here, $T_{s}^{p}$ and $T_{k}^{p}$ represents personalized text prompt, \(\text{concat}(\cdot)\) represents the concatenation operation, \(F_t(\cdot)\) refers to the process of converting text prompt into a sequence of tokens within the VLM, and \(E(\cdot)\) denotes a learnable Visual Guider, which utilizes \textcolor{black}{a IResNet-based face recognition network} 
% an IResNet50-based architecture \cite{duta2021improved} 
to process a reference image, designed to extract individual-specific information. The inclusion of \(I_r\) serves to prevent emotional leakage. Next, we feed the concatenated tokens into the VLM's semantic encoder to generate the personalized text embeddings:
\begin{align}
\begin{split}
T_{s}^{f} &= \text{\(E_t\)}(T_{s}^{p}) \\
T_{k}^{f} &= \text{\(E_t\)}(T_{k}^{p})
\end{split}
\end{align}
where $\text{\(E_t\)}(\cdot)$ represents the semantic encoder in the VLM, which remains fixed during the learning process. For emotion-centric visual embedding, while the VLM's visual encoder can capture emotional cues from images, these cues are not always specific to facial features due to biases in the training data. Various elements, such as colors, can also evoke emotions. To address this, we propose using an Emotion Projector to isolate and emphasize emotion-related information within the visual data:
\begin{align}
\begin{split}
I_{s}^{f} = F_p(\text{\(E_v\)}(I_s))
\end{split}
\end{align}

Among them, $\text{\(E_v\)}(\cdot)$ denotes the fixed visual encoder in the VLM, while \(F_p(\cdot)\) is learnable. To enhance the learning of emotion-centric visual and personalized text embeddings, we leverage the inherent connections between visual and semantic modalities within the VLM. Specifically, embeddings correctly paired with their respective emotions are designated as positive samples, while incorrectly paired embeddings are treated as negative samples. Through cross-modal emotional correlation learning, we maximize the similarity between positive samples and minimize it for negative ones:
\begin{align}
\begin{split}
\mathcal{L}_{1} = (1 - \text{sim}(T_{s}^{f}, I_{s}^{f})) + \text{sim}(T_{k}^{f}, I_{s}^{f})
\end{split}
\label{eq:pepl_loss}
\end{align}
where $\text{sim}(\cdot)$ represents the similarity calculation. Through backpropagation, we can guide $E(\cdot)$ to leverage the visual priors from the VLM's visual encoder to learn information about how a specific individual expresses emotions from visual images. Simultaneously, we can guide $F_p(\cdot)$ to utilize the emotional semantic priors from the VLM's semantic encoder to further capture emotional cues from the visual information.

\subsection{Visual-Text Emotional Differencing Correlation Regularization}
\label{PCMECL:vtedc}
% Once the PEPL module is trained, we can extract aligned personalized text embeddings and emotion-centric visual embeddings from a single facial image and its corresponding emotion prompt in a single pass. To effectively supervise the training of the SPFEM model, a intuitive idea is to leverage PEPL: using the output image from SPFEM to derive emotion-centric visual embeddings and the corresponding target emotion prompt to derive personalized text embeddings. By enforcing consistency between these embeddings, we can guide the generation of visual images that more accurately reflect the target emotion. However, significant distribution discrepancies between the visual and text features arise due to inherent differences between these modalities. Even after training PEPL to align them, these discrepancies persist, leading to imprecise supervision signals. To address this challenge, we propose Visual-Text Emotional Differencing Correlation Regularization (VTEDC). Instead of directly aligning the two modality features, we align the differences between source and target emotion features within the visual modality and those within the semantic modality. This method bridges the inherent gap between modalities, as shown in Fig. \ref{fig: VTEDC}, enabling PEPL to more effectively supervise the training of the SPFEM model.
\textcolor{black}{
Once the PEPL module is pre-trained, it can extract aligned, personalized emotional embeddings from image-text pairs. A straightforward approach to supervise the SPFEM model would be to enforce consistency between the visual features of its output image and the textual features of the target emotion prompt—a strategy we term absolute feature alignment. However, this intuitive idea faces a fundamental challenge: the inherent modality gap. As illustrated in Fig. \ref{fig: CMECC} (a) and substantiated by our high-dimensional feature analysis in Section III of the Supplementary Material, a structural discrepancy exists between the visual and textual feature manifolds. Forcing a direct, absolute alignment between features from these distinct spaces can introduce biases and hinder learning.
}

\textcolor{black}{
To address this, we propose the Visual-Text Emotional Differencing Correlation Regularization (VTEDC), a novel paradigm that focuses on difference alignment instead. This approach offers two key advantages. First, by aligning the relative change from a source to a target emotion (i.e., the difference vector) rather than the absolute states, it cleverly bypasses the need to align absolute positions across disparate modalities. We only require the direction of change to be consistent, which is a more robust and achievable objective. Second, this process implicitly performs feature disentanglement. The visual difference vector, computed between two images of the same individual, naturally cancels out static, emotion-irrelevant factors such as identity and background. This yields a purer representation of the emotional variation, reducing noise and enhancing the precision of the supervision. This makes the supervision signal robust not only to the modality gap but also to intra-subject variations.
}

\textcolor{black}{
Our ablation study in Table \ref{table: ablation_full} provides strong empirical evidence for this paradigm, showing a significant performance drop when VTEDC is removed.
}

\begin{figure}[t]
\centering
\includegraphics[width=0.9\columnwidth]{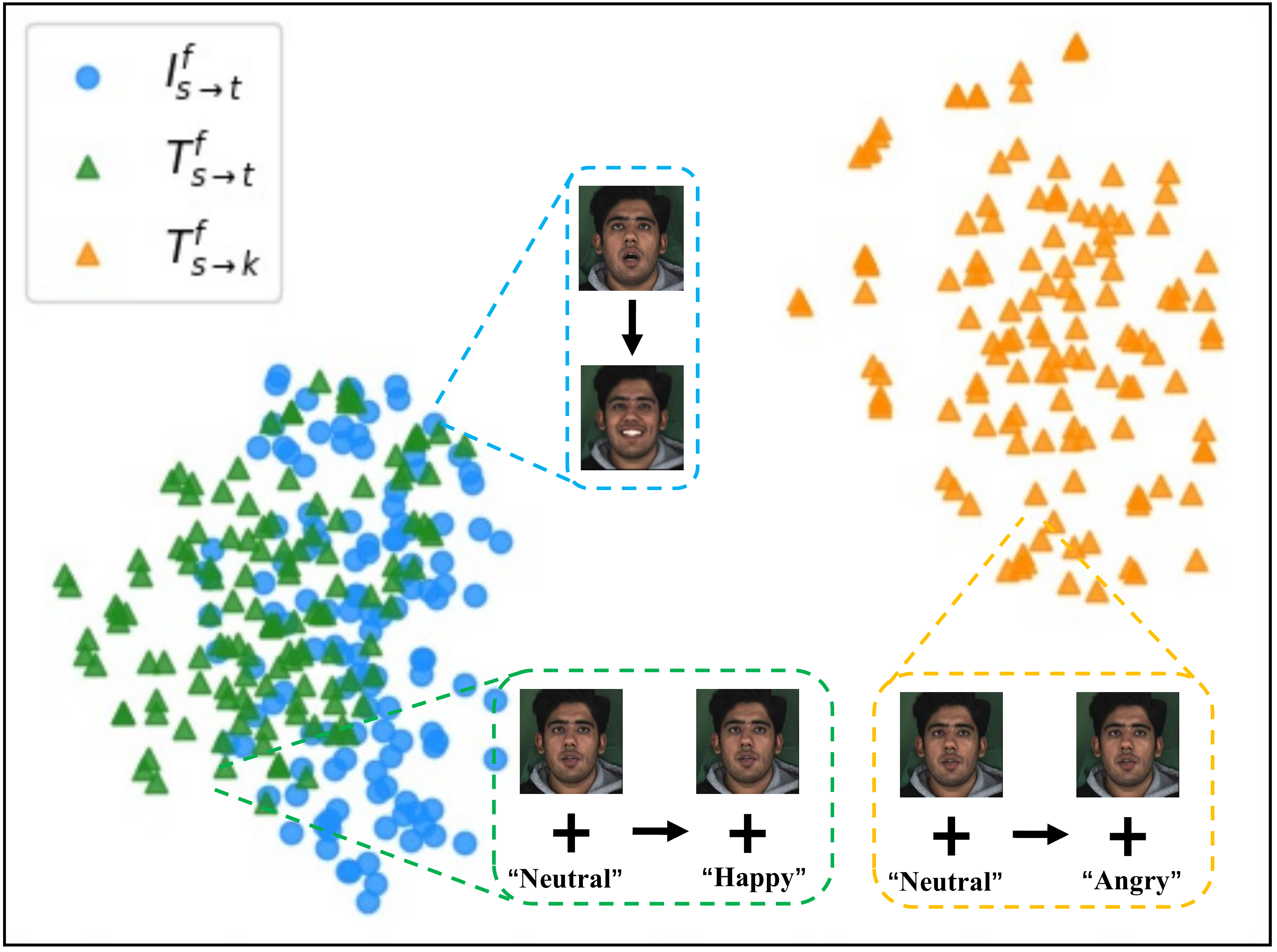} 
% \caption{The distribution of visual and text features disparities. The visual feature disparities are well-aligned with the corresponding text feature disparities, while distinctly separating from non-corresponding text feature disparities.}
\caption{
\textcolor{black}{
    \textbf{A t-SNE visualization of the learned feature difference space, using features extracted from the MEAD dataset.} 
    Each point is a difference vector from a "neutral" source: 
    blue circles (\(I^f_{s \to t}\)) are visual differences; 
    green triangles (\(T^f_{s \to t}\)) are their corresponding text differences; 
    and orange triangles (\(T^f_{s \to k}\)) are non-corresponding text differences (e.g., "neutral"\(\to\)"angry" text vs. "neutral"\(\to\)"happy" image). 
    The tight clustering of corresponding blue/green pairs (e.g., "happy") and their clear separation from non-corresponding orange pairs (e.g., "angry") demonstrates a discriminative and cross-modally consistent feature difference space.
    % visually confirms the effectiveness of our VTEDC mechanism in learning a discriminative, cross-modally consistent space.
}
}
\label{fig: VTEDC}
\end{figure}

Formally, the input image, output image, source emotion prompt, and target emotion prompt of the SPFEM model are denoted as \( I_s \), \( I_t \), \( T_s \), and \( T_t \), respectively. We sequentially feed the visual-text pair \( I_s \) and \( T_s \), and the visual-text pair \( I_t \) and \( T_t \) into the PEPL module to obtain their corresponding personalized text embeddings and emotion-centric visual embeddings:
\begin{align}
\begin{split}
I_{s}^{f}, T_{s}^{f} &= F_{pepl}(I_s, T_s) \\
I_{t}^{f}, T_{t}^{f} &= F_{pepl}(I_t, T_t)
\end{split}
\end{align}

Next, we calculate the difference between \( I_{s}^{f} \) and \( I_{t}^{f} \), and the difference between \( T_{s}^{f} \) and \( T_{t}^{f} \) in the aligned embedding space:
\begin{align}
\begin{split}
I_{s \rightarrow t}^{f} &= I_{s}^{f} - I_{t}^{f} \\
T_{s \rightarrow t}^{f} &= T_{s}^{f} - T_{t}^{f}
\end{split}
\end{align}

We enforce that the difference between \( I_{s}^{f} \) and \( I_{t}^{f} \) in the embedding space aligns with the difference between \( T_{s}^{f} \) and \( T_{t}^{f} \) to guide the SPFEM model in generating visual images that more accurately reflect the target emotion:
\begin{align}
\begin{split}
\mathcal{L}_{2} = 1 - \text{sim}(I_{s \rightarrow t}^{f}, T_{s \rightarrow t}^{f})
\end{split}
\end{align}

\( \mathcal{L}_{2} \) is built upon PEPL and VTEDC. The former enhances the current VLM by incorporating a learnable Visual Guider and an Emotion Projector to obtain personalized text embeddings and emotion-centric visual embeddings. The latter, leveraging the visual-semantic embedding pairs learned through PEPL, further reduces the gap within modalities using the difference technique. Through backpropagation, \( \mathcal{L}_{2} \) serves as a more precise supervisory signal for emotion generation in the SPFEM model.

\textcolor{black}{
\subsection{SPFEM Training with PCMECL Supervision}
}
\label{PCMECL:training_supervision}

\textcolor{black}{
With the PEPL module pre-trained (Section~\ref{PCMECL:pepl}) and the \(\mathcal{L}_{2}\) loss defined (Section~\ref{PCMECL:vtedc}), we now formulate the complete end-to-end training objective for an SPFEM model supervised by our PCMECL framework.
}

\textcolor{black}{
\noindent\textbf{Overall Objective.}
}
\textcolor{black}{
The training of the SPFEM model is guided by a composite loss function, \(L_{\text{total}}\), which combines the baseline's original objective \(L_{\text{SPFEM}}\) with our proposed cross-modal regularization term \(L_{2}\):
}

\begin{equation}
\textcolor{black}{
    L_{\text{total}} = L_{\text{SPFEM}} + \lambda \mathcal{L}_{2}
}
\label{eq:total_loss}
\end{equation}

\textcolor{black}{
where \(L_{\text{SPFEM}}\) (defined in Section~\ref{PCMECL:preliminaries}) ensures the fundamental generation quality, such as identity preservation and lip-synchronization. Our \(\mathcal{L}_{2}\) loss provides the crucial emotional-semantic guidance. The hyperparameter \(\lambda\) balances these two objectives.
}

\textcolor{black}{
\noindent\textbf{Training Pipeline.}
}
\textcolor{black}{
The end-to-end training process proceeds as follows. In each training step, a batch of data is passed through the SPFEM model's generator to produce a set of output images \(I_t\). Subsequently, these generated images \(I_t\), along with the corresponding source images \(I_s\) and text prompts \(T_s, T_t\), are fed into the frozen, pre-trained PEPL module. The PEPL module computes the feature embeddings required to calculate the \(\mathcal{L}_{2}\) loss, as detailed in Section~\ref{PCMECL:vtedc}. Finally, the total loss \(L_{\text{total}}\) is computed, and its gradients are backpropagated to update the parameters of the SPFEM generator. This joint optimization process steers the generator towards producing results that are not only high-fidelity but also emotionally accurate and consistent with the textual guidance.
}

\textcolor{black}{
A key advantage of our framework is its robustness and practicality. Once pre-trained, the PCMECL module operates as a frozen, plug-and-play supervisory block. This paradigm of pre-training the module only once ensures a highly consistent application method that requires no further tuning during the main SPFEM training. As demonstrated in our experiments (Section V), this single frozen module shows consistent effectiveness across different baseline architectures and datasets, validating its robustness.
}

\subsection{Implementation Details}

\noindent\textbf{PEPL Details.} 
% Within the PEPL module, to focus more intently on acquiring information related to emotions during the image feature extraction process, we first use the VLM's visual encoder to extract image features. Subsequently, we introduce an Emotion Projector to further map these features to obtain emotion-related image characteristics. Specifically, to achieve more precise feature mapping, we have trained seven distinct mapping networks corresponding to the seven emotions present in our dataset (i.e., angry, disgusted, fear, happy, neutral, sad, surprised), each dedicated to mapping the features of their respective emotion. Once the PEPL module is trained, it is integrated into the SPFEM model for supervision. During this phase, the emotion of the image input into the PEPL module dictates the use of the corresponding Emotion Projector network to map the image and extract features associated with that particular emotion.
\textcolor{black}{
The PEPL module is composed of two key components: the Visual Guider and the Emotion Projector. Besides, the PEPL module is built upon the pre-trained CLIP-ViT-B/32 model as its VLM backbone.
}

\textcolor{black}{
The Visual Guider employs a hybrid "frozen backbone + learnable head" architecture. The backbone is a pre-trained and frozen IResNet50-based network, while the learnable MLP head is trained jointly with PEPL’s contrastive learning objective. This design leverages powerful priors for robust identity representation with adaptability for personal nuances.
}

\textcolor{black}{
The Emotion Projector is designed to extract fine-grained visual features corresponding to a specific emotion from the source image \(I_s\). To achieve high precision and avoid feature entanglement between different emotions, we implement it as seven distinct and specialized networks \textcolor{black}{corresponding} to the seven emotions in dataset. Each projector is a three-layer MLP with hidden dimensions of [512, 256, 512] and a ReLU activation function. This multi-network design is an engineering trade-off for high fidelity in emotion representation, validated by our ablation study in Section V of the Supplementary Material, as it captures nuanced differences better than a single network.
}

% \textcolor{black}{
% This design choice of using seven specialized projectors, instead of a single conditional network, is a deliberate engineering decision to achieve the highest possible fidelity in emotion representation. We observed that a single network tends to learn coarse and averaged representations, struggling to capture the fine-grained visual cues that differentiate similar emotions. In contrast, specialized projectors allow each network to dedicate its full capacity to modeling the nuances of a single emotion. This rationale is backed by a quantitative ablation study in the Supplementary Material, where the multi-network design achieves a significantly higher CSIM score while keeping FAD and LSE-D scores stable.
% }

\textcolor{black}{
During pre-training, the parameters of the Visual Guider's MLP and Emotion Projectors are jointly optimized via a cross-modal contrastive learning objective, guided by carefully constructed positive and negative pairs.
}

\textcolor{black}{
We construct positive pairs from matching image-text embeddings and negative pairs using a structured sampling strategy that selects semantically distant emotions. This strategy, detailed in Section VI of the Supplementary Material, ensures a robust learning signal. The contrastive learning objective, as described in Eq. (\ref{eq:pepl_loss}), leverages these pairs to maximize positive similarity and minimize negative similarity. Through backpropagation, the model's parameters are updated to learn discriminative features.
}

\textcolor{black}{
Once pre-trained, all learnable parameters within the PEPL module are frozen. The module then combines with VTEDC to provide supervisory signals for the SPFEM model.
}

\noindent\textbf{Training Details.} During the training phase of PEPL module, we leveraged the GeForce RTX 3090 which the amount of memory is 24G, and implemented a Stochastic Gradient Descent (SGD) \cite{robbins1951stochastic} optimizer. We initiated the training with an SGD optimizer learning rate set at 0.1. Strategically, we reduced the learning rate by an order of magnitude, decrementing it by a factor of 10 at the second, fourth, and sixth epochs. This approach was part of a 10-epoch training regimen designed to refine the model's learning efficiency and convergence. Following the training of the PEPL module, we incorporate it with the VTEDC strategy into the SPFEM model's training process for the supervision of emotional manipulation. \textcolor{black}{
In the SPFEM model training process, we fine-tune the key hyperparameter \(\lambda\) in our overall loss function (\(L_{\text{total}}\) in Eq. \ref{eq:total_loss}) balances the baseline loss and our PCMECL supervision. Based on a sensitivity analysis detailed in Section IV of the Supplementary Material, we found that different baselines require different optimal values. For all our experiments, we set (\(\lambda_{\text{NED}}\) = 0.4) for the NED baseline and (\(\lambda_{\text{ICface}}\) = 0.05) for the ICface baseline. These values were chosen as they provided the best trade-off between FAD, LSE-D and CSIM on our validation set.
}

\section{Experiments}
\subsection{Experiment Setting}
\noindent\textbf{Datasets.} MEAD \cite{wang2020mead} comprises 60 speakers, which each speaker records 30 videos in each emotional state (i.e., angry, disgusted, fear, happy, neutral, sad, surprised). For the training of our PEPL module, we selected videos from 36 speakers that contain 7560 videos. To evaluate its performance, we incorporate it into SPFEM models, NED and ICface. And chose 6 distinct speakers (M003, M009, M012, M030, W015 and W029) that contain 1260 videos to train SPFEM models. We randomly assigned 90\% of these video for training and the remaining 10\% for testing. 
\textcolor{black}{
Furthermore, to evaluate the generalization capability of our framework, we conducted experiments on the RAVDESS dataset~\cite{livingstone2018ryerson}. Crucially, these experiments were performed in a zero-shot, cross-dataset setting: the entire model, including both the SPFEM baseline (NED/ICface) and our PCMECL module, was trained exclusively on MEAD and then directly tested on RAVDESS without any retraining or fine-tuning.
}
Specifically, we also chose 6 speakers (actors 1-6), including 168 videos. 

\noindent\textbf{Evaluation Metrics.} We demonstrate the performance of our method through a comprehensive evaluation on multiple metrics, which have been widely used in previous research. We employ these metrics for evaluation: \textbf{ 1) Frechet Arcface Distance (FAD)} \cite{heusel2017gans} evaluate realism of video by calculating  the divergence between the feature vectors of generated and authentic videos, utilizing current advanced face recognition technology \cite{deng2019arcface}. A lower FAD score signifies a higher degree of realism; \textbf{ 2) Lip Sync Error Distance (LSE-D)} \cite{prajwal2020lip} gauges the precision of lip-audio synchronization, employing a pre-trained model \cite{chung2017out} to quantify the discrepancies between lip movements and audio tracks. A lower LSE-D value corresponds to higher lip-audio synchronization; \textbf{ 3) Consine similarity (CSIM)} \cite{zhao2021robust} evaluate emotional similarity by comparing similarity of the embeddings extracted from a state-of-the-art expression recognition network in both generated and real videos. A higher CSIM score indicates a greater level of emotional similarity. Our results are represent under two settings: inter-identification, where emotion reference and source video share the same speaker, and cross-identification, involving difference speakers. \textcolor{black}{For a more rigorous treatment of these metrics, including their precise mathematical formulations and step-by-step calculation procedures, please refer to Section II of the Supplementary Materials.}

\begin{table*}[!t]
\centering
\caption{\textcolor{black}{Comparison results of average FAD, LSE-D and CSIM of NED, ICface, and SSERD with and without our PCMECL in the Inter-ID and Cross-ID settings on the MEAD and RAVDESS datasets. Detailed per-emotion results are provided in Section IX of the Supplementary Material.}}
\label{table:sota_main_comparison}
\scriptsize
\renewcommand{\arraystretch}{1.2}
\setlength{\tabcolsep}{15pt}

\begin{tabular}{l|l|ccc|ccc}
\toprule
\multirow{2}{*}{\textbf{Datasets}} & \multirow{2}{*}{\textbf{Methods}} & \multicolumn{3}{c|}{\textbf{Inter-ID}} & \multicolumn{3}{c}{\textbf{Cross-ID}} \\
\cline{3-8}
& & FAD $\downarrow$ & LSE-D $\downarrow$ & CSIM $\uparrow$ & FAD $\downarrow$ & LSE-D $\downarrow$ & CSIM $\uparrow$ \\
\hline 

\multirow{6}{*}{\textbf{MEAD}} 
& NED & 2.108 & 9.454 & 0.831 & 4.448 & 9.906 & 0.773 \\
& \textbf{Ours(NED)} & \textbf{1.075} & \textbf{9.364} & \textbf{0.915} & \textbf{4.300} & \textbf{9.374} & \textbf{0.784} \\
\cline{2-8}
& ICface & 6.795 & 10.083 & 0.775 & 9.540 & 11.238 & 0.688 \\
& \textbf{Ours(ICface)} & \textbf{6.604} & \textbf{9.566} & \textbf{0.794} & \textbf{9.384} & \textbf{10.261} & \textbf{0.711} \\
\cline{2-8}
& SSERD & 0.740 & 9.126 & 0.904 & 2.453 & 9.200 & 0.848 \\
    & \textbf{Ours(SSERD)} & \textbf{0.655} & \textbf{9.038} & \textbf{0.908} & \textbf{2.441} & \textbf{9.093} & \textbf{0.857} \\

\hline 

\multirow{6}{*}{\textbf{RAVDESS}} 
& NED & 3.057 & 7.562 & 0.825 & 5.412 & 8.034 & 0.760 \\
& \textbf{Ours(NED)} & \textbf{2.736} & \textbf{7.460} & \textbf{0.833} & \textbf{5.204} & \textbf{7.933} & \textbf{0.772} \\
\cline{2-8}
& ICface & 8.443 & 8.480 & 0.755 & 9.424 & 11.539 & 0.677 \\
& \textbf{Ours(ICface)} & \textbf{7.972} & \textbf{8.286} & \textbf{0.778} & \textbf{9.207} & \textbf{10.582} & \textbf{0.681} \\
\cline{2-8}
& SSERD & 1.399 & 7.441 & 0.894 & 3.360 & 7.621 & 0.790 \\
& \textbf{Ours(SSERD)} & \textbf{1.296} & \textbf{7.355} & \textbf{0.899} & \textbf{3.360} & \textbf{7.519} & \textbf{0.801} \\
\bottomrule
\end{tabular}
\end{table*}

\begin{figure}[t]
\centering
% \vspace{-10pt}
\includegraphics[width=\columnwidth]{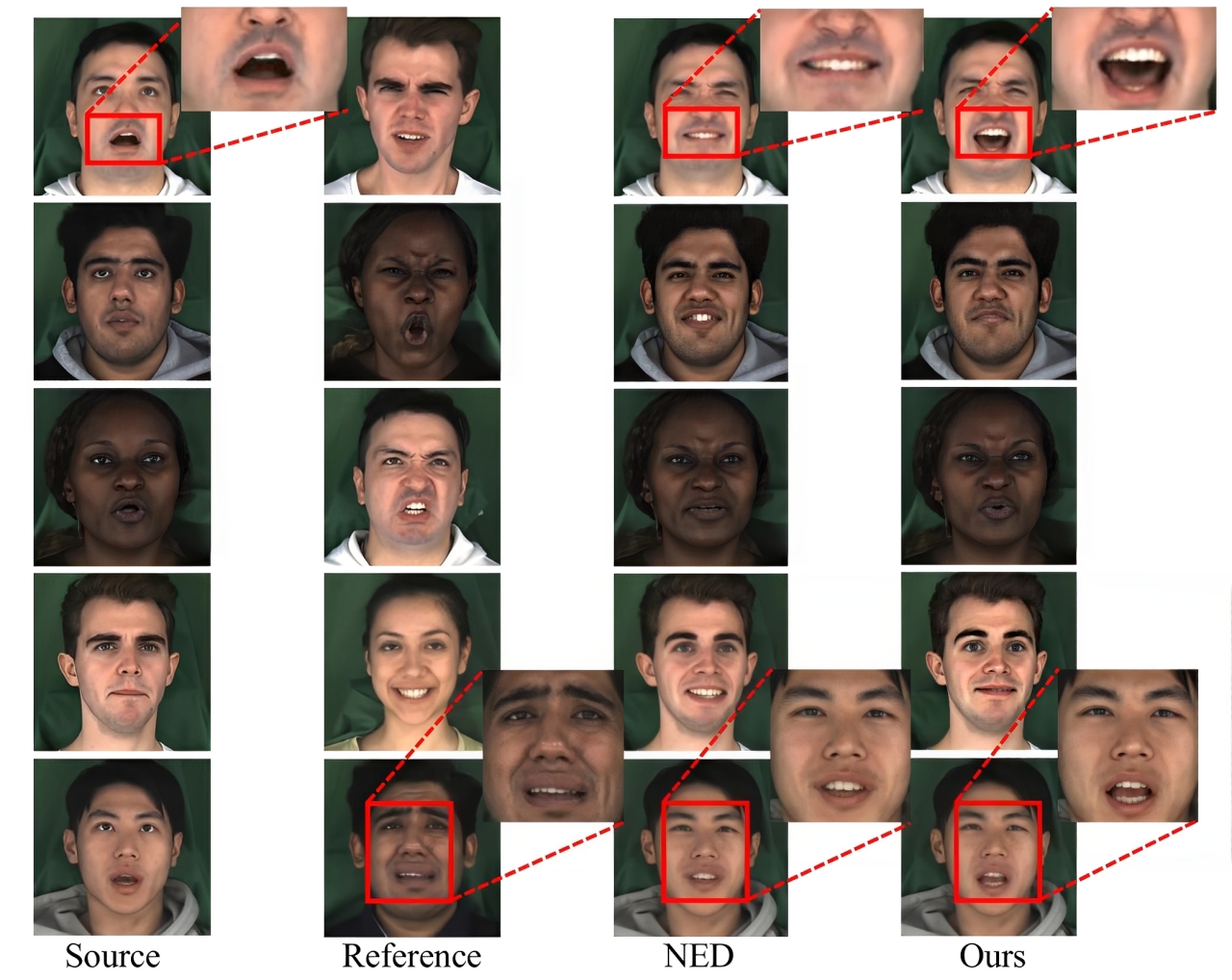} 
\caption{
\textcolor{black}{
Qualitative comparisons of NED with and without PCMECL supervision on the MEAD dataset.
% We transfer the emotion from a Reference image to a Source identity. The `Ours' column shows the final results from the PCMECL-supervised NED. Red boxes and their corresponding zoom-in panels highlight our method's superior detail in key expressive regions.
}
}
\label{fig: NED_MEAD_comparison}
\end{figure}

\begin{figure}[t]
\centering
\includegraphics[width=\columnwidth]{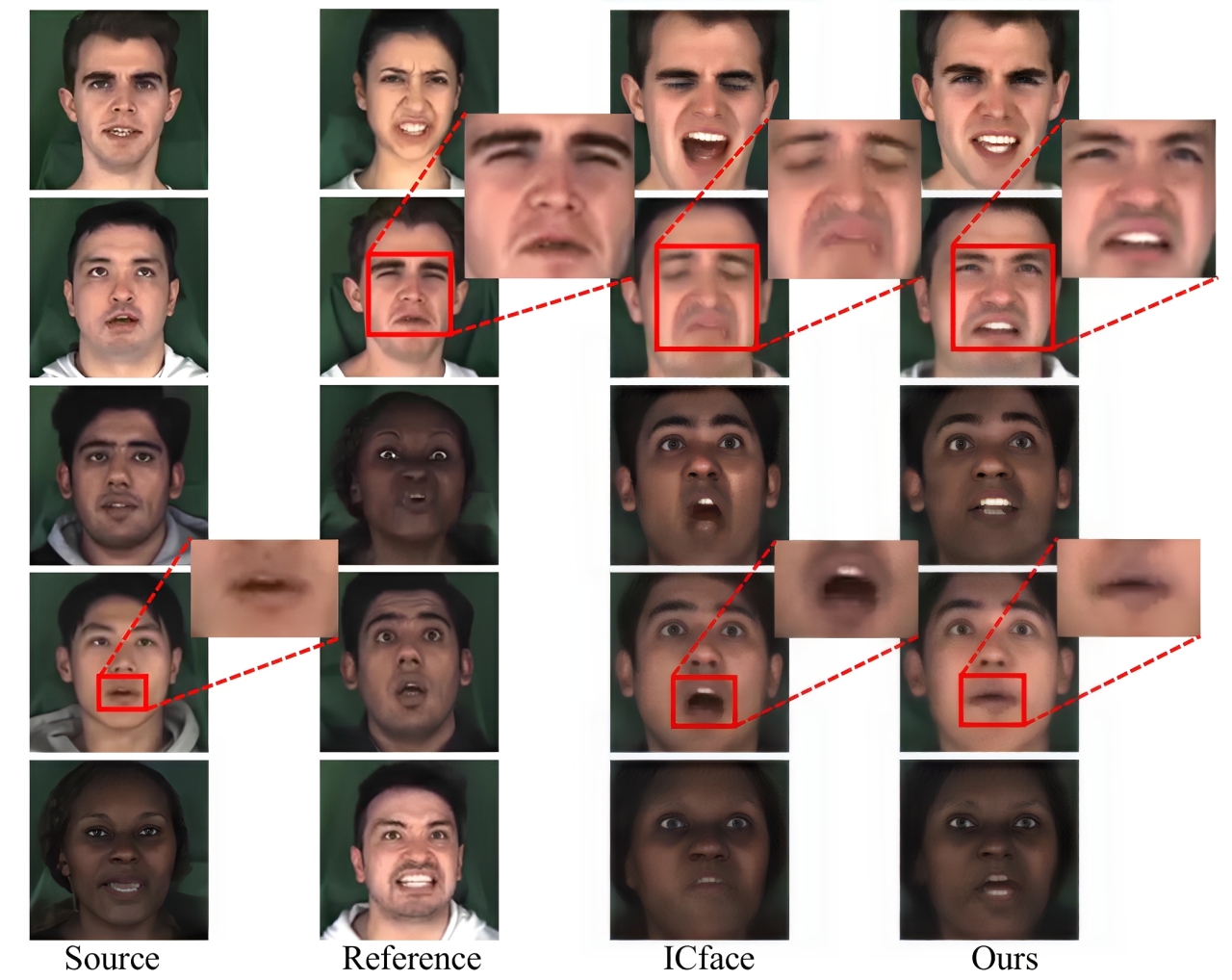} 
\caption{
\textcolor{black}{
Qualitative comparisons of ICface with and without PCMECL supervision on the MEAD dataset.
% We transfer the emotion from a Reference image to a Source identity. The `Ours' column shows the final results from the PCMECL-supervised ICface. Red boxes and their corresponding zoom-in panels highlight our method's superior detail in key expressive regions.
}
}
% \vspace{-5pt}
\label{fig: ICface_MEAD_comparison}
\end{figure}

\begin{figure}[t]
\centering
\includegraphics[width=\columnwidth]{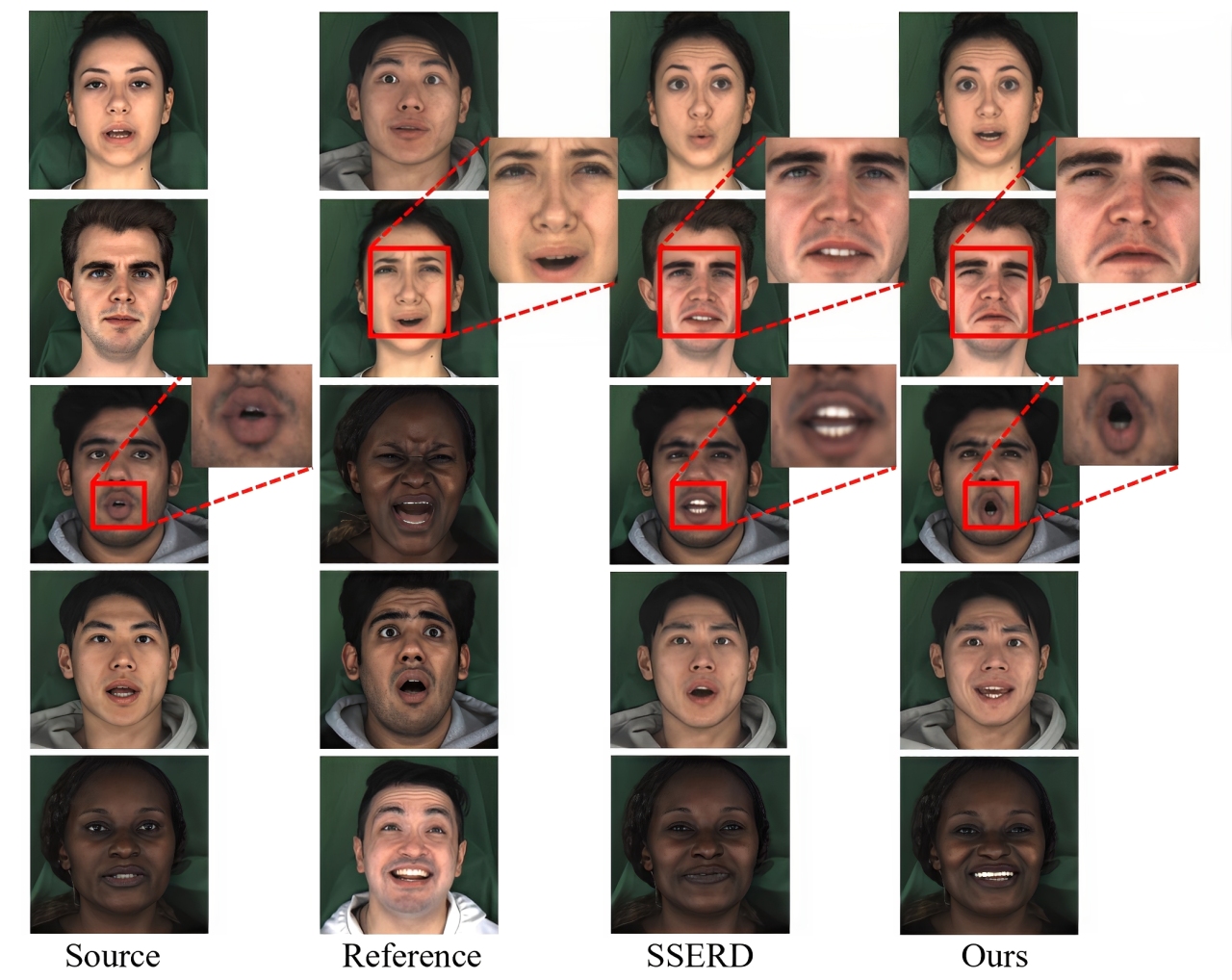} 
\caption{
\textcolor{black}{
Qualitative comparisons of SSERD with and without PCMECL supervision on the MEAD dataset.
% We transfer the emotion from a Reference image to a Source identity. The `Ours' column shows the final results from the PCMECL-supervised SSERD. Red boxes and their corresponding zoom-in panels highlight our method's superior detail in key expressive regions.
}
}
\label{fig: SSERD_MEAD_comparison}
\end{figure}

\subsection{Quantitative Comparisons}
\textcolor{black}{
We begin by presenting the comprehensive performance comparisons on the MEAD dataset in Table \ref{table:sota_main_comparison}. Our framework demonstrates consistent improvements across distinct architectures, from the two-stage method (NED) and the single-stage method (ICface) to the latest state-of-the-art method (SSERD).
}

\textcolor{black}{
\noindent\textbf{Performance on the MEAD Dataset.} Integrating PCMECL into NED significantly enhances the resulting image sequences. In the Inter-ID setting, there is a notable improvement over the NED baseline: FAD decreases from 2.108 to 1.075, LSE-D drops from 9.454 to 9.364, and CSIM increases from 0.831 to 0.915. Among these metrics, CSIM shows particularly significant improvements, attributed to PCMECL's crucial role in supervising the training via the PEPL module and VTEDC regularization. Specifically, PCMECL extracts emotion-centric visual embeddings and corresponding personalized text embeddings through the PEPL module. By using VTEDC to bridge the gap between these two modalities and ensure their alignment, PCMECL guides the generation of visual images that more accurately reflect the target emotion. Additionally, under this supervision, the SPFEM model is capable of generating more realistic facial expressions, resulting in a generated data distribution that is more in line with real data, thereby leading to the noticeable reduction in FAD scores. Similarly, integrating PCMECL into the single-stage ICface method results in significant gains. In the Inter-ID setting, PCMECL improves ICface's performance with reductions in FAD (6.795 to 6.604) and LSE-D (10.083 to 9.566), and increases in CSIM (0.775 to 0.794).
}

\textcolor{black}{
Most importantly, we validate our framework on the state-of-the-art method, SSERD. As shown in Table \ref{table:sota_main_comparison}, although SSERD achieves impressive baselines using StyleGAN latent space, incorporating PCMECL pushes the boundaries further. Ours(SSERD) achieves the best overall performance, reducing FAD to 0.655 and raising CSIM to 0.908 in the Inter-ID setting. This indicates that while SSERD relies on synthetic paired data which may contain semantic noise, PCMECL introduces direct VLM-based supervision to enforce stricter alignment with the target emotional concept, yielding the highest fidelity and emotional accuracy.
}

\textcolor{black}{
\noindent\textbf{Zero-shot Generalization on the RAVDESS Dataset.} 
To demonstrate the robustness of our framework, we conducted a strict zero-shot evaluation on the RAVDESS dataset, where models trained exclusively on MEAD were directly tested without fine-tuning.
As reported in the bottom section of Table \ref{table:sota_main_comparison}, incorporating PCMECL yields consistent improvements across all architectures even in this challenging setting.
When applied to ICface and NED, our method significantly reduces FAD and LSE-D while improving CSIM in both Inter-ID and Cross-ID settings.
Notably, for the state-of-the-art baseline SSERD, our method further enhances its zero-shot performance, reducing FAD from 1.399 to 1.296 and increasing CSIM from 0.894 to 0.899 in the Inter-ID setting. These consistent improvements across varied methods and data domains underscore the robustness of PCMECL as a plug-and-play supervisory module.
}

\subsection{Qualitative Comparisons}
\textcolor{black}{
In this section, we showcase the visualization results of the representative baselines (NED, ICface) and the state-of-the-art method (SSERD) on the MEAD dataset. We compare the outcomes with and without the application of our PCMECL algorithm, as illustrated in Fig. \ref{fig: NED_MEAD_comparison}, \ref{fig: ICface_MEAD_comparison}, and \ref{fig: SSERD_MEAD_comparison}. In these visualizations, we transfer the emotion from a Reference image to a Source identity. The `Ours' column displays the final results supervised by PCMECL, where red boxes and corresponding zoom-in panels are provided to highlight the superior detail in key expressive regions. Additional comparisons on the RAVDESS dataset are provided in Section X of the Supplementary Material.
}

\textcolor{black}{
\noindent\textbf{1) Emotional Similarity:} 
The third columns of the figures reveal distinct behaviors across methods. Representative baselines (NED, ICface) often produce inaccurate or ambiguous expressions. This limitation stems from their failure to provide guidance specific to individuals, which frequently leads to unnatural emotion editing and visual artifacts. While the SOTA method SSERD generates high-quality faces, it relies on synthetic paired data for supervision. This can lead to semantic misalignments, where the generated expression is plausible but fails to fully capture the intensity or specific nuance of the Reference emotion. In contrast, the fourth columns demonstrate that our PCMECL framework, supervised by the PEPL module's personalized prompts, generates expressions that are semantically much closer to the target. This confirms PCMECL's ability to provide fine-grained semantic guidance beyond simple geometric deformation or synthetic priors.
}

\textcolor{black}{
\noindent\textbf{2) Realism:} 
For NED and ICface (Fig. \ref{fig: NED_MEAD_comparison} and \ref{fig: ICface_MEAD_comparison}), the baseline results often exhibit significant visual artifacts, such as facial distortions and unnatural blurs, attributed to destructive edits caused by models struggling to match vague or flawed emotional objectives. For instance, NED's inaccurate 3DMM predictions and ICface's difficulty in decoupling attributes lead to such distortions. In the case of SSERD (Fig. \ref{fig: SSERD_MEAD_comparison}), although the image quality is inherently high due to the StyleGAN backbone, it often suffers from identity texture smoothing or slight identity drift due to the GAN inversion bottleneck. Our method mitigates this by enforcing consistency in the semantic feature difference space via VTEDC. As shown in the zoom-in panels, PCMECL better preserves fine-grained identity details while achieving the desired emotion, leading to visually more natural and faithful results across all architectures.
}

\textcolor{black}{
\noindent\textbf{3) Lip-Audio Preservation Accuracy:} 
Consistent mouth movement is critical for SPFEM. Baselines often exhibit inconsistent lip shapes when forcing an emotional change. This is rooted in the trade-off between expression intensity and lip-sync constraints. Our PCMECL focuses on aligning feature differences rather than enforcing absolute states, allowing the model to learn emotional changes without distorting the mouth shape required by the audio. As shown in the zoom-in panels, our method preserves the original mouth movements more accurately, ensuring high lip-sync fidelity even under strong emotional transformations.
}

\begin{table*}[!t]
\vspace{-2pt}
\caption{Realism, emotion similarity, and mouth shape similarity ratings of the user study on NED with and without PCMECL, and on ICface with and without PCMECL on MEAD dataset.}
\label{table:user_study_MEAD}
\vspace{-5pt}
\scriptsize
\centering
\begin{tabular}{c|cccc|cccc|cccc}
\toprule
\multirow{2}{*}{\textbf{Emotion}} & \multicolumn{4}{c|}{\textbf{Realism}} & \multicolumn{4}{c|}{
    \textbf{\makecell[c]{Emotion similarity}}} & \multicolumn{4}{c}{\textbf{\makecell[c]{Mouth shape similarity}}} \\
\cline{2-13}
& \textbf{NED} & \textbf{Ours(NED)} & \textbf{ICface} & \textbf{Ours(ICface)} & \textbf{NED} & \textbf{Ours(NED)} & \textbf{ICface} & \textbf{Ours(ICface)} & \textbf{NED} & \textbf{Ours(NED)} & \textbf{ICface} & \textbf{Ours(ICface)} \\
\hline
Neutral & 32\% & 68\% & 55\% & 45\% & 48\% & 52\% & 45\% & 55\% & 40\% & 60\% & 40\% & 60\% \\
Angry & 28\% & 72\% & 45\% & 55\% & 28\% & 72\% & 32\% & 68\% & 31\% & 69\% & 34\% & 66\% \\
Disgusted & 44\% & 56\% & 42\% & 58\% & 40\% & 60\% & 38\% & 62\% & 30\% & 70\% & 38\% & 62\% \\
Fear & 36\% & 64\% & 48\% & 52\% & 46\% & 54\% & 46\% & 54\% & 36\% & 64\% & 42\% & 58\% \\
Happy & 28\% & 72\% & 38\% & 62\% & 36\% & 64\% & 40\% & 60\% & 28\% & 72\% & 32\% & 68\% \\
Sad & 36\% & 64\% & 35\% & 65\% & 28\% & 72\% & 35\% & 65\% & 26\% & 74\% & 38\% & 62\% \\
Surprised & 48\% & 52\% & 48\% & 52\% & 40\% & 60\% & 42\% & 58\% & 33\% & 67\% & 36\% & 64\% \\
\hline 
\textbf{Avg.} & \textbf{36\%}& \textbf{64\%}& \textbf{44\%}& \textbf{56\%}& \textbf{38\%}& \textbf{62\%}& \textbf{40\%}& \textbf{60\%}& \textbf{32\%}& \textbf{68\%}& \textbf{37\%}& \textbf{63\%} \\
\bottomrule
\end{tabular}
\end{table*}

\begin{table*}[!t]
\vspace{-2pt}
\caption{Realism, emotion similarity, and mouth shape similarity ratings of the user study on NED with and without PCMECL, and on ICface with and without PCMECL on RAVDESS dataset.}
\label{table:user_study_RAVDESS}
\vspace{-5pt}
\scriptsize
\centering
\begin{tabular}{c|cccc|cccc|cccc}
\toprule
\multirow{2}{*}{\textbf{Emotion}} & \multicolumn{4}{c|}{\textbf{Realism}} & \multicolumn{4}{c|}{
    \textbf{\makecell[c]{Emotion similarity}}} & \multicolumn{4}{c}{\textbf{\makecell[c]{Mouth shape similarity}}} \\
\cline{2-13}
& \textbf{NED} & \textbf{Ours(NED)} & \textbf{ICface} & \textbf{Ours(ICface)} & \textbf{NED} & \textbf{Ours(NED)} & \textbf{ICface} & \textbf{Ours(ICface)} & \textbf{NED} & \textbf{Ours(NED)} & \textbf{ICface} & \textbf{Ours(ICface)} \\
\hline
Neutral & 36\% & 64\% & 54\% & 46\% & 45\% & 55\% & 48\% & 52\% & 36\% & 64\% & 45\% & 55\% \\
Angry & 30\% & 70\% & 38\% & 62\% & 38\% & 62\% & 38\% & 62\% & 40\% & 60\% & 32\% & 68\% \\
Disgusted & 50\% & 50\% & 40\% & 60\% & 40\% & 60\% & 50\% & 50\% & 32\% & 68\% & 42\% & 58\% \\
Fear & 35\% & 65\% & 44\% & 56\% & 50\% & 50\% & 48\% & 52\% & 44\% & 56\% & 44\% & 56\% \\
Happy & 32\% & 68\% & 35\% & 65\% & 35\% & 65\% & 40\% & 60\% & 35\% & 65\% & 38\% & 62\% \\
Sad & 46\% & 54\% & 35\% & 65\% & 38\% & 62\% & 38\% & 62\% & 38\% & 62\% & 36\% & 64\% \\
Surprised & 52\% & 48\% & 46\% & 54\% & 45\% & 55\% & 50\% & 50\% & 40\% & 60\% & 40\% & 60\% \\
\hline 
\textbf{Avg.} & \textbf{40\%}& \textbf{60\%}& \textbf{42\%}& \textbf{58\%}& \textbf{42\%}& \textbf{58\%}& \textbf{45\%}& \textbf{55\%}& \textbf{38\%}& \textbf{62\%}& \textbf{40\%}& \textbf{60\%} \\
\bottomrule
\end{tabular}
\end{table*}

\subsection{User study}
We executed online user studies to evaluate the comparative performance of the NED and ICface, both with and without the PCMECL algorithm applied. The study was structured into three key segments: realism, emotion similarity, and mouth shape similarity, across seven fundamental emotions. For each emotion, we select 10 videos from MEAD dataset, amassing a collection of 70 videos. In the study, 25 participants were enrolled to assess three key dimensions of every video. As outlined in Table \ref{table:user_study_MEAD}, incorporating PCMECL on MEAD dataset, the NED and ICface method outperformed their standard versions across realism, emotion similarity, and mouth shape similarity. On average, the integration of the PCMECL led to notable enhancements, with a 28\% increase in realism, a 24\% increase in emotion similarity, and a substantial 36\% increase in mouth shape similarity when compared to the NED baseline. Simultaneously, the ICface method, augmented with PCMECL, has seen improvements across all metrics. On average, the integration of PCMECL into the ICface baseline has yielded substantial improvements: a 12\% increase in realism, a 20\% boost in emotion similarity, and a 26\% enhancement in mouth shape similarity.

Additionally, Table \ref{table:user_study_RAVDESS} showcases the results derived from utilizing the NED and ICface baselines with the RAVDESS dataset. Considering the limited video selection within RAVDESS, we meticulously picked 5 videos per emotion, amassing a compilation of 35 videos. This curated set was subsequently appraised by an identical group of 25 participants. Our research reveals that the adoption of the \textcolor{black}{PCMECL} algorithm consistently garners substantially elevated ratings across the three key dimensions assessed within the RAVDESS dataset.

\subsection{Ablation Study}
\textcolor{black}{
To validate that our proposed PEPL and VTEDC modules are necessary advancements over both an original baseline and a standard VLM-based supervision approach, we conducted a comprehensive ablation study. We compare five key settings to demonstrate the step-by-step improvements from our contributions. The results, evaluated on the NED baseline under both Inter-ID and Cross-ID settings, are summarized in Table \ref{table: ablation_full}.
}

\subsubsection{Effectiveness of VLM-based Supervision}
\textcolor{black}{
To validate our central claim that leveraging VLMs is a beneficial direction for SPFEM, we first compare the performance of the original baseline (a) NED against a (b) Standard VLM (SVLM) supervision baseline and our (e) full model. The SVLM baseline directly aligns the visual features of the generated image with the semantic features from a generic text prompt (e.g., "a photo of a happy face").
}

\textcolor{black}{
As shown in Table \ref{table: ablation_full}, the comparison between (a) NED and (b) SVLM demonstrates the potential of leveraging VLMs, with SVLM showing improvements in FAD and CSIM in the Inter-ID setting. However, the performance gain is limited, and even shows a slight degradation in some metrics (e.g., LSE-D). This confirms our hypothesis that a naive application is insufficient. In stark contrast, the substantial leap in performance from (b) SVLM to (e) Ours across all metrics, especially the significant CSIM increase from 0.848 to 0.915 in the Inter-ID setting, provides direct, compelling evidence for the necessity and effectiveness of our specialized PEPL and VTEDC modules.
}

\subsubsection{Analyses of VTEDC} 
\textcolor{black}{
Building upon the established benefits of VLM-based supervision, we now dissect the distinct contributions of our proposed VTEDC modules.
}
VTEDC can eliminate the inherent modality gap in the VLM embedding space by using difference-based techniques. We verify this by comparing it with another baseline that directly constrains the SPFEM output image's embedding to be consistent with its corresponding emotion text embedding (referred to as 'Ours w/o VTEDC'). 
\textcolor{black}{As rows (c) and (e) in Table \ref{table: ablation_full} show,} our approach outperforms NED across all metrics. When we integrate VTEDC into the supervision process, we achieve  more optimized results. This validates that VTEDC can effectively eliminate differences between modalities and provides better supervision for training the SPFEM model.

\begin{figure}[t]
\centering
\includegraphics[width=0.9\columnwidth]{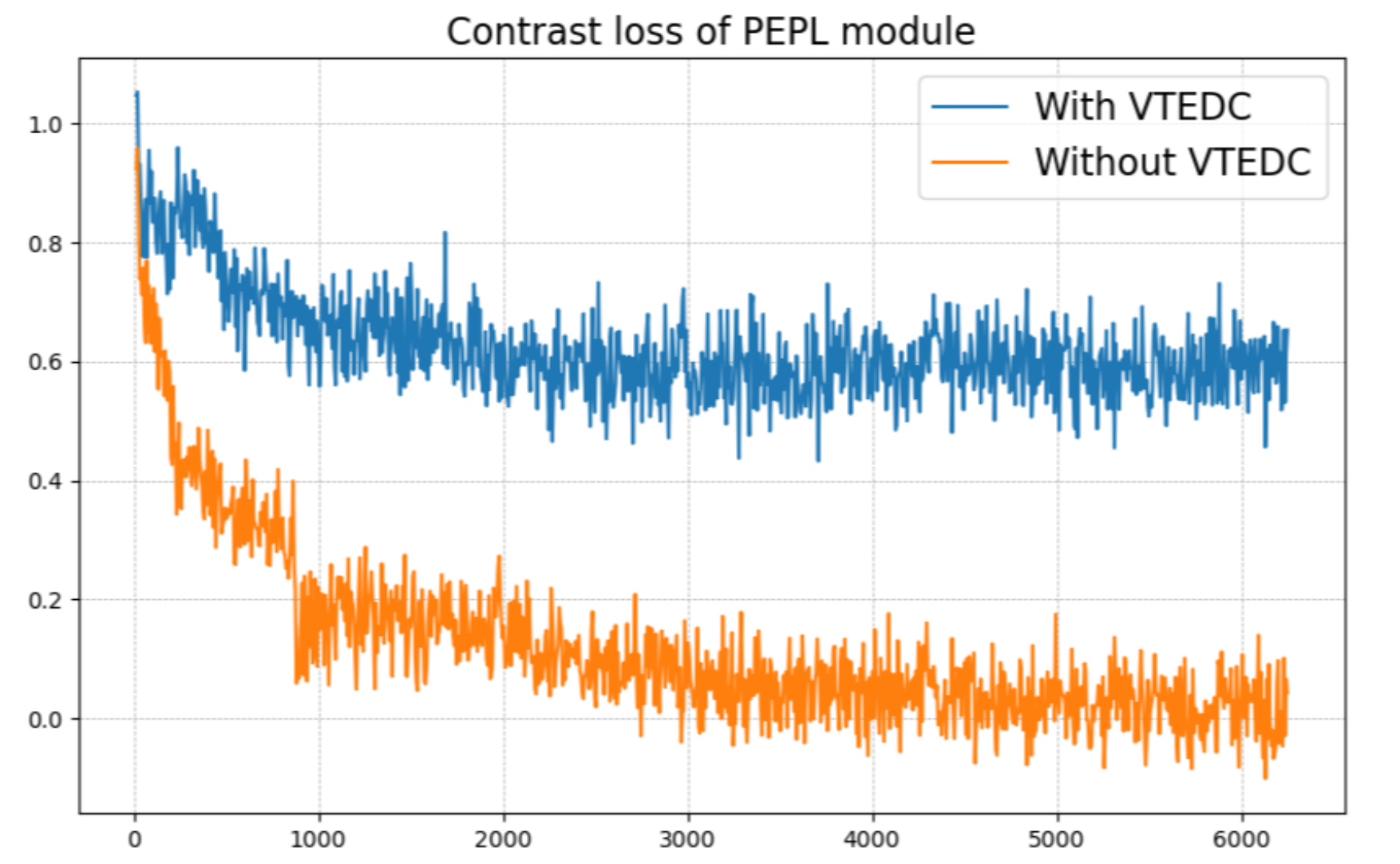} 
\caption{Convergence curve of the PEPL module training with or without VTEDC strategy. (Blue: with VTEDC; Orange: without VTEDC)}
% \vspace{-10pt}
\label{fig: EITRL_Module_Training_Loss}
\end{figure}

Another intuitive approach is to directly integrate VTEDC into the PEPL module's learning process. During PEPL training, the difference technique can be employed to eliminate the modal gap, allowing the learned PEPL module to be seamlessly integrated into SPFEM. This integration ensures that the output image's embedding aligns with its corresponding emotion text embedding, thereby providing effective supervision. However, applying the difference technique directly during PEPL training presents challenges in achieving convergence, as shown in Fig. \ref{fig: EITRL_Module_Training_Loss}. The orange curve illustrates that the PEPL module, when trained without VTEDC, achieves complete convergence. In contrast, the blue curve indicates that incorporating VTEDC during PEPL training results in less satisfactory convergence, with the loss stabilizing at around 0.6. This issue may arise because such training does not align well with VLM's original pre-training approach, thereby failing to fully leverage VLM's prior knowledge. Consequently, we incorporate VTEDC into the supervision process of SPFEM, utilizing the aligned, fine-grained visual-text embedding pairs learned by PEPL, combined with VTEDC, to offer more accurate supervision.

\subsubsection{Analyses of PEPL} 
Since each individual expresses the same emotion in a unique way, the PEPL module incorporates individual appearance to learn personalized emotional prompts, capturing the expressive variations specific to each person. 
We validate this by comparing it with a variant that removes the Visual Guider (referred to as 'Ours w/o VG') and thus relies on non-personalized, generic text prompts for supervision, isolating the impact of our personalization strategy. \textcolor{black}{As indicated in rows (d) and (e) of Table \ref{table: ablation_full},} removing individual appearance results in a deterioration across all three metrics, particularly CSIM and FAD. In the inter-id setting, CSIM drops from 0.915 to 0.852, and FAD increases from 1.075 to 1.812. This decline is due to the lack of consideration for individual differences in expressing the same emotion during supervision, leading to suboptimal performance in CSIM. Additionally, the convergence of data distribution between generated and real expressions contributes to the improvement of FAD. Furthermore, due to inaccuracies in expressions and reduced generation quality, LSE-D is also affected. This demonstrates the crucial role of individual appearance in supervising the SPFEM process.

\begin{table}[h]
    \centering
    \caption{\textcolor{black}{FAD, LSE-D, and CSIM of NED baseline, SVLM (Standard VLM supervision), Ours w/o VTEDC, Ours w/o VG (Visual Guider), and Ours.}}
    \label{table: ablation_full}
    \begin{tabular}{l|l|ccc}
        \toprule
        \textbf{Settings} & \textbf{Methods} & \textbf{FAD}$\downarrow$ & \textbf{LSE-D}$\downarrow$ & \textbf{CSIM}$\uparrow$ \\
        \hline
        \multirow{5}{*}{Inter-ID} 
        & (a) NED                   & 2.108 & 9.454 & 0.831 \\
        & (b) SVLM                  & 1.915 & 9.503 & 0.848 \\
        & (c) Ours w/o VTEDC        & 1.143 & 9.452 & 0.910 \\
        & (d) Ours w/o VG           & 1.812 & 9.526 & 0.852 \\
        & (e) Ours                  & \textbf{1.075} & \textbf{9.364} & \textbf{0.915} \\
        \hline
        \multirow{5}{*}{Cross-ID} 
        & (a) NED                   & 4.448 & 9.906 & 0.773 \\
        & (b) SVLM                  & 4.439 & 9.782 & 0.775 \\
        & (c) Ours w/o VTEDC        & 4.435 & 9.466 & 0.775 \\
        & (d) Ours w/o VG           & 4.477 & 9.591 & 0.777 \\
        & (e) Ours                  & \textbf{4.300} & \textbf{9.374} & \textbf{0.784} \\
        \bottomrule
    \end{tabular}
\end{table}

% \begin{table}[!t]
% \vspace{-2pt}
% \caption{FAD, LSE-D, and CSIM of Ours, Ours w/o VTEDC, and NED baseline.}
% \label{table: Ablation-CMECC}
% \vspace{-5pt}
% \centering
% \small
% \begin{tabular}{c|c|ccc}
% \toprule
% \textbf{Settings} & \textbf{Methods} & \textbf{FAD}$\downarrow$ & \textbf{LSE-D}$\downarrow$ & \textbf{CSIM}$\uparrow$ \\
% \hline
% \multirow{4}{*}{\textbf{Inter-ID}} & NED & 2.108 & 9.454 & 0.831 \\
% & Ours w/o VTEDC  & 1.143& 9.452& 0.910\\
% & \textbf{Ours} & \textbf{1.075}& \textbf{9.364}& \textbf{0.915}\\
% \hline
% \multirow{4}{*}{\textbf{Cross-ID}} & NED & 4.448 & 9.906 & 0.773 \\
% & Ours w/o VTEDC & 4.435& 9.466& 0.775\\
% & \textbf{Ours} & \textbf{4.300}& \textbf{9.374}& \textbf{0.784}\\
% \bottomrule
% \end{tabular}
% \end{table}

% \begin{table}[!t]
% \vspace{-2pt}
% \caption{FAD, LSE-D, and CSIM of Ours, Ours w/o VG (Visual Guider), and NED baseline.}
% \label{table: Ablation-Visual_Guider}
% \vspace{-5pt}
% \centering
% \small
% \begin{tabular}{c|c|ccc}
% \toprule
% \textbf{Settings} & \textbf{Methods} & \textbf{FAD}$\downarrow$ & \textbf{LSE-D}$\downarrow$ & \textbf{CSIM}$\uparrow$ \\
% \hline
% \multirow{4}{*}{\textbf{Inter-ID}} & NED & 2.108 & 9.454 & 0.831 \\
% % & Ours w/o Visual Guider  & 1.812& 9.526& 0.852\\
% & Ours w/o VG  & 1.812& 9.526& 0.852\\
% & \textbf{Ours} & \textbf{1.075} & \textbf{9.364} & \textbf{0.915} \\
% \hline
% \multirow{4}{*}{\textbf{Cross-ID}} & NED & 4.448 & 9.906 & 0.773 \\
% % & Ours w/o Visual Guider & 4.477& 9.591& 0.777\\
% & Ours w/o VG & 4.477& 9.591& 0.777\\
% & \textbf{Ours} & \textbf{4.300}& \textbf{9.374}& \textbf{0.784}\\
% \bottomrule
% \end{tabular}
% \end{table}

\section{Limitations}
\label{sec: limitations}
\textcolor{black}{
While our experiments demonstrate PCMECL's broad effectiveness, its performance is context-dependent. The consistently more marginal gains on the ICface compared to the NED model highlight the representational bottlenecks of baselines. Furthermore, the supervisory precision is limited by the VLM's inherent ability to discern subtle emotions like "Fear," resulting in more moderate gains. Finally, metric sensitivities on low-variance tasks like "Neutral" can lead to performance fluctuations, such as the slight FAD degradation observed in some cases. A detailed analysis is provided in Section VII of the Supplementary Material.
}

\textcolor{black}{
In addition to these performance-related factors, there are two key design limitations. First, in line with standard benchmarks in this field (e.g., MEAD, RAVDESS), our framework currently focuses on seven foundational emotions. Consequently, its ability to generate more complex or nuanced emotional states has not yet been explored. Second, the current hyperparameter selection relies on a manual search. Although this method ensures stability, its efficiency remains a limitation for rapid deployment scenarios.
}

\section{Conclusion}
\textcolor{black}{
In this paper, we introduced PCMECL, a novel plug-and-play supervisory framework that addresses the critical lack of fine-grained, personalized emotional supervision in SPFEM tasks. We were the first to systematically identify and tackle the dual challenges of the modality gap and lack of personalization in VLM-based supervision. Our solution, featuring the PEPL module with its dynamic textual inversion and the VTEDC module with its innovative difference alignment paradigm, was shown to significantly enhance the emotional accuracy and overall quality of state-of-the-art SPFEM models.
}

\textcolor{black}{
Based on the limitations discussed above, we believe promising avenues for future research include the development of co-designed SPFEM and VLM architectures for more synergistic supervision.
\textcolor{black}{
Another key direction is the extension of our framework to generate more complex, nuanced emotions. Specifically, exploring flexible architectures, such as a Mixture-of-Emotions (MoE) model, offers a promising pathway to represent and manipulate emotions within a continuous or blended space. Finally, improving the practicality of our method through automated hyperparameter tuning presents another important direction for future exploration. 
}
We believe PCMECL provides a solid foundation for these future explorations in controllable and personalized audio-visual synthesis.
}

\clearpage
\twocolumn

\setcounter{section}{0}
\renewcommand{\thesection}{S\arabic{section}}
\renewcommand{\thesubsection}{S\arabic{section}.\arabic{subsection}}
\renewcommand{\thefigure}{S\arabic{figure}}
\renewcommand{\thetable}{S\arabic{table}}
\renewcommand{\theequation}{S\arabic{equation}}

\setcounter{figure}{0}
\setcounter{table}{0}
\setcounter{equation}{0}

% supplementary_material.tex
% Do not include \documentclass, \usepackage, \title, \author, or \maketitle here.

\clearpage

\begin{center}
{\LARGE\bfseries Supplementary Material\par}
\end{center}
\vspace{1em}
% \appendices

\section{Loss Functions of the Baseline Models}
\label{section:baseline_losses}

% This section provides a detailed description of two original loss functions of the baseline models, both of which are denoted as \(L_{\text{SPFEM}}\) in our framework.

This section provides a detailed description of the original loss functions for our two baseline models, NED and ICface. In our main paper, both of these are generically referred to as \(L_{\text{SPFEM}}\) and are combined with our proposed supervisory term, \(L_{\text{PCMECL}}\), for end-to-end training. Understanding these baseline objectives is crucial for contextualizing how our framework enhances their performance.

\subsection{Loss Function of NED}
The training of the NED model~\cite{papantoniou2022neural} is guided by a composite loss function, which is a weighted sum of four key objectives:
\begin{equation}
    L_{\text{NED}} = \lambda_{\text{adv}} L_{\text{adv}} + \lambda_{\text{sty}} L_{\text{sty}} + \lambda_{\text{cyc}} L_{\text{cyc}} + \lambda_{\text{sp}} L_{\text{sp}}
\end{equation}
Here, \(L_{\text{adv}}\) is an adversarial loss to ensure the generated expression sequences are realistic. The style reconstruction loss, \(L_{\text{sty}}\), forces the output sequence's style to match the one extracted by the style encoder. The cycle consistency loss, \(L_{\text{cyc}}\), encourages the generator to preserve the content of the input sequence by reconstructing the original style vector. Finally, to ensure audio-visual synchronization, the speech-preserving loss, \(L_{\text{sp}}\), explicitly constrains the mouth-related parameters to remain unchanged during the emotion translation process. The \(\lambda\) terms are the corresponding weighting factors. This comprehensive loss function ensures the baseline model generates identity-preserving and well-synchronized results, upon which our PCMECL is applied to enhance emotional accuracy.

\subsection{Loss Function of ICface}
The ICface model~\cite{tripathy2020icface} is trained with a different composite objective function, expressed as:
\begin{equation}
    L_{\text{ICface}} = L_{\text{adv}}(G_N, D) + L_{\text{adv}}(G_A, D) + \lambda_1 L_{\text{FA}} + \lambda_2 L_{\text{I}} + \lambda_3 L_{\text{R}}
\end{equation}
Here, the total adversarial objective consists of two separate losses, \(L_{\text{adv}}(G_N, D)\) and \(L_{\text{adv}}(G_A, D)\), which are applied to the neutral-face generator (\(G_N\)) and the reenactment generator (\(G_A\)) respectively, to ensure photorealism. A key innovation in ICface is the facial attribute reconstruction loss, \(L_{\text{FA}}\), which extends the discriminator's role to also regress facial attributes from the generated images. The identity classification loss, \(L_{\text{I}}\), employs a pre-trained LightCNN network to ensure the preservation of the subject's identity. Finally, the reconstruction loss, \(L_{\text{R}}\), is an L1 pixel-wise loss that enforces accurate reconstruction of the target frame. The \(\lambda\) terms are the weighting factors for the non-adversarial. 
\textcolor{black}{
\subsection{Loss Function of SSERD}
The SSERD model~\cite{xu2024self} employs a hybrid training strategy that leverages both synthetic paired data and real unpaired data to ensure photorealism and accurate emotion transfer. The overall objective function $\mathcal{L}_{all}$ is defined as:
\begin{equation}
    \mathcal{L}_{all} = \mathbb{I}(I^s)[\mathcal{L}_{pix} + \alpha \mathcal{L}_{lpips}] + \lambda_1 \mathcal{L}_{adv} + \lambda_2 \mathcal{L}_{cl}
\end{equation}
Here, $\mathbb{I}(I^s)$ is an indicator function that returns 1 for synthetic images and 0 for real images. For synthetic paired data, the model is supervised by the pixel-wise L2 loss ($\mathcal{L}_{pix}$) and the VGG-based perceptual loss ($\mathcal{L}_{lpips}$) to enforce content consistency. For all data, a generative adversarial loss ($\mathcal{L}_{adv}$) is applied to enhance realism, and a contrastive loss ($\mathcal{L}_{cl}$) operates on the latent code to strictly disentangle emotion representations from other attributes. The balance coefficients are set to $\alpha=0.8$, $\lambda_1=0.1$, and $\lambda_2=0.1$.
}

When our PCMECL framework is applied, the respective \(L_{\text{SPFEM}}\) for each baseline is combined with our \(L_{\text{PCMECL}}\) term for end-to-end training.

\section{Detailed Evaluation Metrics}
\label{section:metrics}

% This section provides a detailed description of the evaluation metrics used in our study to ensure full clarity, transparency, and reproducibility.

This section provides a comprehensive breakdown of the evaluation metrics used in our study to ensure full clarity, transparency, and reproducibility. We evaluate our method's performance across three key dimensions: the realism and identity preservation of the generated faces (FAD), the accuracy of audio-lip synchronization (LSE-D), and the fidelity of the emotional expression (CSIM). The following subsections detail the calculation principles for each metric, their value ranges, and the criteria for interpreting improvements.

\subsection{Metric Definitions and Calculation Principles}

\paragraph{Fréchet Arcface Distance (FAD)}
FAD is a metric used to evaluate the quality and realism of generated faces, with a strong focus on identity preservation. It is a variant of the widely-used Fréchet Inception Distance (FID)~\cite{heusel2017gans}, but it replaces the general-purpose InceptionV3 network with a pre-trained ArcFace~\cite{deng2019arcface} model, which is specialized for face recognition. The calculation proceeds as follows: 1)~A large set of real images (\(X_r\)) and generated images (\(X_g\)) are collected. 2)~Each image is passed through the pre-trained ArcFace model to extract a 512-dimensional feature embedding. 3)~The collections of embeddings are modeled as two multivariate Gaussian distributions, and their means (\(\mu_r, \mu_g\)) and covariance matrices (\(\Sigma_r, \Sigma_g\)) are calculated. 4)~The FAD score is the Fréchet distance between these two distributions:
\begin{equation}
    \text{FAD} = ||\mu_r - \mu_g||^2 + \text{Tr}(\Sigma_r + \Sigma_g - 2(\Sigma_r \Sigma_g)^{1/2})
\end{equation}

\paragraph{Lip Sync Error Distance (LSE-D)}
LSE-D measures the synchronization between the lip movements in a video and the corresponding audio track. We follow the methodology established by state-of-the-art lip-sync models like Wav2Lip~\cite{prajwal2020lip}. The process is as follows: 1)~For a given video segment, the audio waveform and the corresponding mouth region-of-interest (ROI) video frames are extracted. 2)~Both are fed into a pre-trained, powerful SyncNet~\cite{chung2016lip} model. 3)~The SyncNet model outputs an audio embedding (\(E_a\)) and a visual embedding (\(E_v\)) for each short time window. 4)~The LSE-D is the average Euclidean distance between these corresponding embeddings over the entire video duration \(T\):
\begin{equation}
    \text{LSE-D} = \frac{1}{T} \sum_{t=1}^{T} || E_a(t) - E_v(t) ||_2
\end{equation}

\paragraph{Cosine Similarity (CSIM)}
CSIM evaluates the emotional accuracy of the generated facial expressions by comparing them to ground-truth videos expressing the same emotion. The calculation is as follows: 1)~For a generated video and its corresponding ground-truth real video, frames are sampled. 2)~Each frame is passed through a pre-trained, state-of-the-art facial expression recognition (FER) network~\cite{zhao2021robust}. 3)~The FER network outputs a high-dimensional feature embedding for each frame. 4)~CSIM is the average cosine similarity between the feature embeddings of the \(N\) pairs of generated frames (\(v_g\)) and real frames (\(v_r\)):
\begin{equation}
    \text{CSIM} = \frac{1}{N} \sum_{i=1}^{N} \frac{v_{g,i} \cdot v_{r,i}}{||v_{g,i}|| \cdot ||v_{r,i}||}
\end{equation}

\subsection{Value Ranges and Interpretation}

To aid in the interpretation of our quantitative results, the value ranges and criteria for each metric are clarified below:
\begin{itemize}
    \item \textbf{FAD \& LSE-D:} These metrics range from \([0, +\infty)\). Lower values are better, indicating higher realism/identity preservation and better lip-sync, respectively.
    \item \textbf{CSIM:} This metric ranges from \([-1, 1]\). Higher values are better, indicating stronger emotional similarity to the ground truth.
\end{itemize}

\subsection{Criteria for "Significant" and "Marginal" Improvements}

To provide a clear and data-driven context for interpreting the magnitude of our results, we establish the following guidelines for what constitutes a meaningful improvement in our task domain.

\paragraph{Quantitative Guideline}
Based on our empirical observations and common practices in the facial generation literature, we consider a relative improvement of 5\% or more on key metrics (e.g., CSIM increasing from 0.800 to 0.840) to be significant. Improvements below 2\% are considered marginal, representing subtle but positive changes.

\paragraph{Qualitative Justification from User Studies}
This quantitative threshold is not arbitrary but is strongly corroborated by our user study, as analyzed in the main paper (see Section V-D). The cases where our method achieved "notable" and "substantial" enhancements, with measured improvements often far exceeding our 5\% significance threshold (e.g., gains of 12\% to 36\% reported in the user study section), are the same ones where our method achieved decisive user preference rates (e.g., a 64\% preference in realism for `Ours(NED)` in Table III of the main paper). This direct correlation confirms that our definition of "significant" translates to perceptually meaningful and impactful results from a human evaluation perspective.

\section{High-Dimensional Feature Analysis}
\label{section: feature analysis}

% In our main paper, we identify a fundamental challenge in applying VLMs for supervision: the inherent modality gap between visual and textual feature spaces. While this is illustrated in 2D visualizations (Fig. 1 (a)), this section provides rigorous quantitative evidence for this phenomenon directly in the original 512-dimensional CLIP feature space, confirming that it is not merely an artifact of dimensionality reduction.

This section provides rigorous quantitative evidence for the "modality gap"—a fundamental challenge in applying VLMs for supervision that we identify in our main paper. While 2D visualizations (Fig. 2 (a)) in the main text illustrate this gap, the analysis here confirms its existence directly in the original 512-dimensional CLIP feature space. Our findings demonstrate that this phenomenon is not merely an artifact of dimensionality reduction and provide a solid, high-dimensional foundation for the design of our VTEDC module.

\subsection{Experimental Setup}
We randomly sampled approximately 7000 images from the MEAD dataset, covering all seven emotion categories. For each image, we extracted its 512-D visual feature embedding using the pre-trained CLIP-ViT-B/32 image encoder. For each of the seven emotion categories, we extracted the corresponding 512-D text feature embedding from its unique text prompt "a photo of a happy face".

\subsection{Metrics Definition}
To quantify the modality gap, we defined two metrics. Our hypothesis is that if a significant gap exists, the cross-modal similarity between corresponding pairs will be substantially lower than the inter-modal similarity among visual examples of the same class. The metrics are:
\begin{itemize}
    \item \textbf{Inter-Modal Cohesion for Images (\(S_{\text{image}}\)):} This measures the visual coherence of an emotion class. It is the average pair-wise cosine similarity between all image feature embeddings within that class.
    
    \item \textbf{Image-Text Matching Similarity (\(S_{\text{match}}\)):} This measures the alignment between the visual features of an emotion and its textual representation. It is the average cosine similarity between all image feature embeddings of an emotion and the single text feature embedding for that emotion.
\end{itemize}

\subsection{Results and Conclusion}
The results of our analysis are presented in Table~\ref{tab:modality_gap}. As shown, for every emotion category, \(S_{\text{match}}\) is significantly lower than \(S_{\text{image}}\). On average, we observe a similarity drop of approximately 0.18 when moving from inter-modal to cross-modal comparisons, which is a substantial difference. 

This quantitatively demonstrates that features from different modalities are inherently more distant from each other than features within the same modality, even for correctly corresponding pairs. This provides a solid, non-visual, high-dimensional foundation for the design of our VTEDC module, which is specifically engineered to bypass this structural discrepancy.

\begin{table}[h]
    \centering
    \caption{
    \textcolor{black}{Quantitative evidence of the modality gap in the 512-D CLIP feature space. All values are average cosine similarities.}
    }
    \label{tab:modality_gap}
    \begin{tabular}{l | c c c}
        \toprule
        \textbf{Emotion Category} & \textbf{\(S_{\text{image}}\)} & \textbf{\(S_{\text{match}}\)} & \textbf{Gap (\(S_{\text{image}} - S_{\text{match}}\))} \\
        \hline
        Angry     & 0.821 & 0.452 & 0.369 \\
        Disgusted & 0.805 & 0.431 & 0.374 \\
        Fear      & 0.853 & 0.485 & 0.368 \\
        Happy     & 0.862 & 0.553 & 0.309 \\
        Neutral   & 0.915 & 0.601 & 0.314 \\
        Sad       & 0.884 & 0.524 & 0.360 \\
        Surprised & 0.849 & 0.538 & 0.311 \\
        \hline
        \textbf{Average} & \textbf{0.856} & \textbf{0.512} & \textbf{0.344} \\
        \bottomrule
    \end{tabular}
\end{table}

\section{Hyperparameter Sensitivity Analysis for \(\lambda\)}
\label{section:hyperparameter}

% The hyperparameter \(\lambda\) in our overall loss function, \(L_{\text{total}} = L_{\text{SPFEM}} + \lambda L_{\text{PCMECL}}\) (where\(L_{\text{PCMECL}}\) is the \( \mathcal{L}_{2} \) is the \( \mathcal{L}_{2} \) proposed in the Section IV of the main paper), critically balances the baseline's generation quality with our proposed cross-modal emotional supervision. To identify the optimal value for \(\lambda\), we conducted a systematic coarse-to-fine search for each baseline model on a held-out validation set. The comprehensive results of this search are summarized in the figures below.

The hyperparameter \(\lambda\) in our overall loss function, \(L_{\text{total}} = L_{\text{SPFEM}} + \lambda L_{\text{PCMECL}}\), critically balances the baseline's generation quality with our proposed cross-modal emotional supervision. 
% The \(L_{\text{PCMECL}}\) is the \( \mathcal{L}_{2} \) term corresponds to the \( \mathcal{L}_{2} \) loss introduced in Section IV of the main paper. 
\textcolor{black}{
The term \(L_{\text{PCMECL}}\) corresponds to the \( \mathcal{L}_{2} \) loss introduced in Section IV of the main paper. 
}
To identify the optimal value for \(\lambda\) for each baseline, we conducted a systematic coarse-to-fine search on a held-out validation set. The following subsections present the comprehensive results of this analysis.
 
\subsection{Sensitivity Analysis for NED (\(\lambda_{\text{NED}}\))}

Our search for the optimal \(\lambda_{\text{NED}}\) proceeded in two stages.

\paragraph{Coarse-grained Search}
We first conducted a search across different orders of magnitude (\(\lambda_{\text{NED}} \in \{0.01, 0.1, 1.0, 10.0\}\)). This initial analysis revealed that a value of \(\lambda=0.01\) provided insufficient emotional guidance, while values of 1.0 or higher led to a significant degradation in generation quality (FAD). This indicated that the optimal value likely resided within the (0.1, 1.0) range.

\begin{figure}[h!]
    \centering
    \includegraphics[width=\columnwidth]{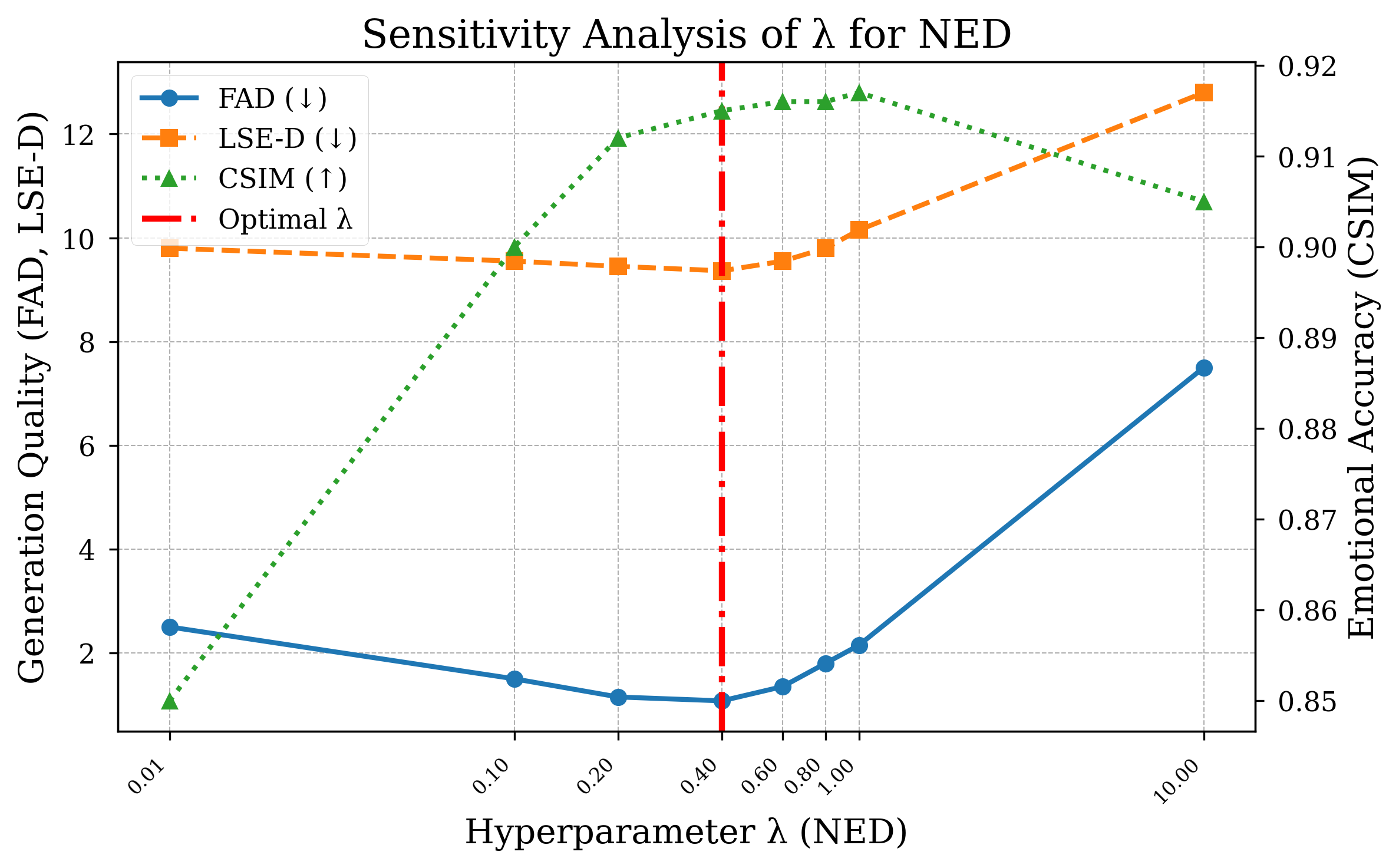}
    \caption{
    \textcolor{black}{Sensitivity analysis of \(\lambda\) for the NED baseline, summarizing our coarse-to-fine search. The plot shows the trade-off between Generation Quality (FAD, LSE-D) and Emotional Accuracy (CSIM). The red line marks our optimal choice.}
    }
    \label{fig:lambda_ned_analysis}
\end{figure}

\paragraph{Fine-grained Search}
Subsequently, we performed a fine-grained linear search within this promising range. The results, illustrated in Figure~\ref{fig:lambda_ned_analysis}, reveal a clear trade-off. As \(\lambda_{\text{NED}}\) increases from 0.2 to 0.4, a substantial improvement in emotional accuracy (CSIM) is achieved, while the generation quality metrics (FAD and LSE-D) remain at their optimal levels. Beyond \(\lambda=0.4\), the gains in CSIM saturate, whereas the quality costs (FAD, LSE-D) begin to rise sharply.

Therefore, we selected \(\lambda_{\text{NED}} = 0.4\) as the optimal value, as it represents the point of optimal trade-off: it achieves near-maximal emotional accuracy (CSIM) just before the generation quality (FAD and LSE-D) begins to significantly degrade.

\subsection{Sensitivity Analysis for ICface (\(\lambda_{\text{ICface}}\))}
A similar two-stage search was performed for the ICface baseline.
\paragraph{Coarse-grained Search}
Given the sensitive adversarial dynamics of the GAN-based ICface model, our coarse-grained search over \(\lambda_{\text{ICface}} \in \{0.01, 0.1, 1.0\}\) confirmed the model's high sensitivity to our supervisory loss. Values of 0.1 or higher were observed to destabilize training and severely degrade generation quality. This confirmed that the optimal value must reside within the narrow (0, 0.1) range.

\begin{figure}[h!]
    \centering
    \includegraphics[width=\columnwidth]{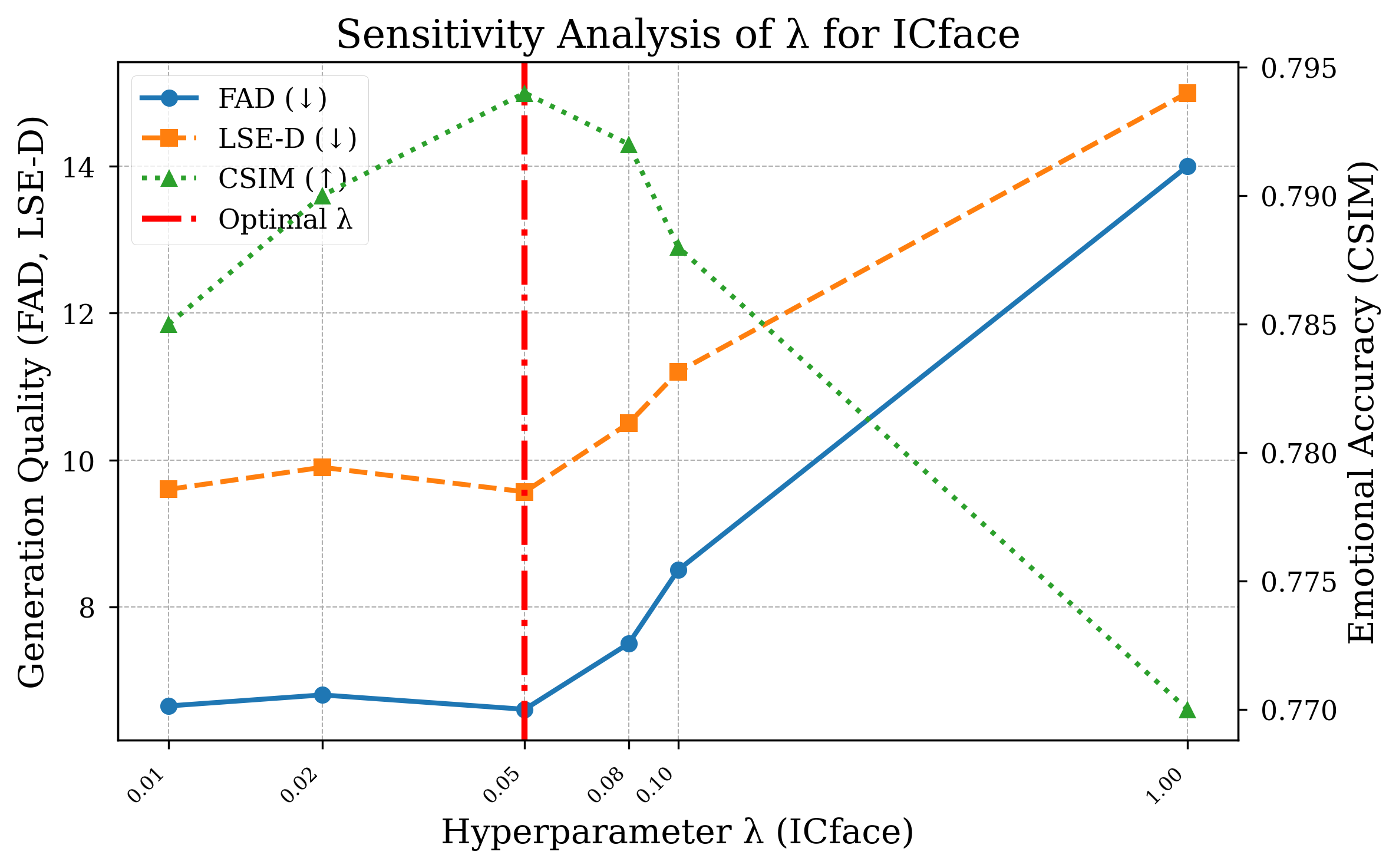}
    \caption{
    \textcolor{black}{Sensitivity analysis of \(\lambda\) for the ICface baseline, summarizing our coarse-to-fine search. The plot shows the trade-off between Generation Quality (FAD, LSE-D) and Emotional Accuracy (CSIM). The red line marks our optimal choice.}
    }
    \label{fig:lambda_icface_analysis}
\end{figure}

\paragraph{Fine-grained Search}
We then conducted a fine-grained search within this range, with the results shown in Figure~\ref{fig:lambda_icface_analysis}. The analysis reveals that \(\mathbf{\lambda_{\text{ICface}} = 0.05}\) provides the peak emotional accuracy (CSIM) while concurrently maintaining the lowest generation quality costs. Values beyond 0.05 show a decline in CSIM and a sharp increase in FAD and LSE-D. Thus, we selected 0.05 as the robust and optimal choice.

\subsection{\textcolor{black}{Sensitivity Analysis for SSERD (\(\lambda_{\text{SSERD}}\))}}
\begin{figure}[h!]
    \centering
    \includegraphics[width=\columnwidth]{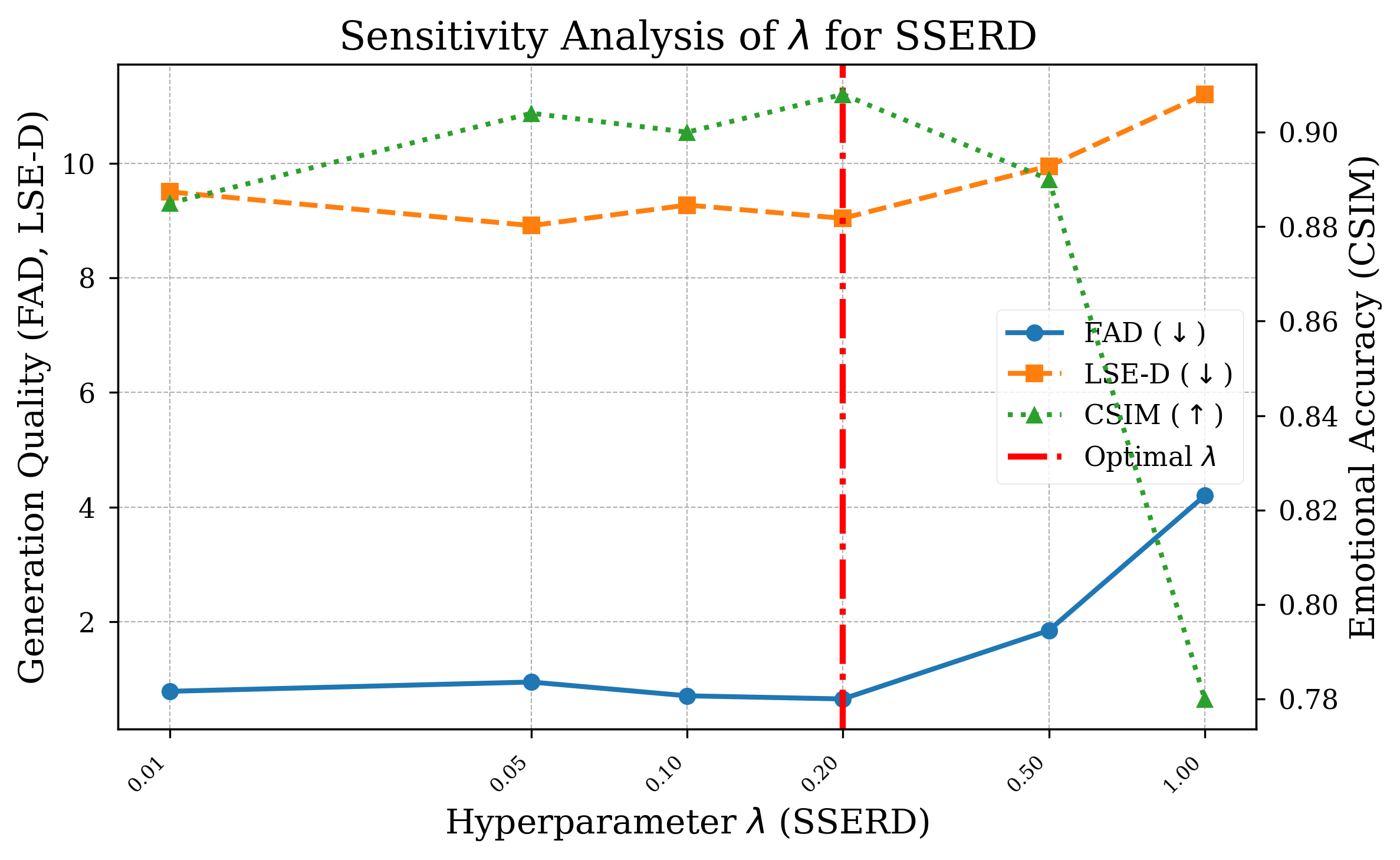}
    \caption{
    \textcolor{black}{Sensitivity analysis of \(\lambda\) for the SSERD baseline, summarizing our coarse-to-fine search. The plot shows the trade-off between Generation Quality (FAD, LSE-D) and Emotional Accuracy (CSIM). The red line marks our optimal choice.}
    }
    \label{fig:lambda_sserd_analysis}
\end{figure}

\textcolor{black}{
We applied the same rigorous two-stage search strategy to the SSERD baseline.
}
\textcolor{black}{   
\paragraph{Coarse-grained Search}
SSERD relies on disentangling emotion representations within the StyleGAN latent spac. An initial coarse-grained search was conducted over \(\lambda_{\text{SSERD}} \in \{0.01, 0.1, 1.0\}\) to identify the optimal parameter range. Specifically, \(\lambda=1.0\) resulted in a noticeable drop in identity preservation (FAD), likely because excessive regularization disrupted the delicate manifold of the pre-trained StyleGAN. Conversely, \(\lambda=0.01\) yielded negligible improvements in emotional alignment. This narrowed our search to the range of (0.1, 1.0).
}
\textcolor{black}{  
\paragraph{Fine-grained Search}
We subsequently performed a fine-grained search within this interval. As illustrated in Figure~\ref{fig:lambda_sserd_analysis}, the performance trends show a distinct peak. Increasing \(\lambda_{\text{SSERD}}\) initially boosts the CSIM score significantly, correcting the semantic misalignments inherent in SSERD's synthetic supervision. However, pushing \(\lambda\) beyond 0.2 leads to diminishing returns in emotional accuracy and a gradual deterioration in image quality metrics (FAD). Consequently, we selected \(\lambda_{\text{SSERD}} = 0.2\) as the optimal setting, where the model achieves the best balance between high-fidelity generation and precise emotional control.
}

\section{Ablation Study on Emotion Projector Design}
\label{section:ablation_projector}

This study validates our design choice of using multiple specialized Emotion Projector networks over a single conditional network. This design choice is a deliberate engineering decision to achieve the highest possible fidelity in emotion representation. We observed that a single network tends to learn coarse and averaged representations, struggling to capture the fine-grained visual cues that differentiate similar emotions. In contrast, specialized projectors allow each network to dedicate its full capacity to modeling the nuances of a single emotion. This ablation study provides the rigorous quantitative evidence.

\subsection{Experimental Setup}
We compare our proposed Multi-Network Design against a Single Conditional Network baseline. The single network is also a three-layer MLP with a similar architecture to one of our specialized projectors, but it takes an additional 7-dimensional one-hot emotion vector as a conditional input. Both PEPL module variants were pre-trained under identical settings on the MEAD dataset and their effectiveness as supervisory signals was evaluated by applying them to the NED baseline.

\subsection{Results and Analysis}
The results of this comparison are presented in Table~\ref{tab:ablation_projector_full}. The data clearly shows that the Multi-Network design achieves a higher score on the emotion similarity (CSIM) metric across both Inter-ID and Cross-ID settings. This substantial gain in our primary objective (emotional accuracy) is achieved while keeping the fundamental generation quality metrics (FAD and LSE-D) largely stable or even slightly improved. This validates our engineering decision to use specialized projectors, as the performance gain in our core task significantly outweighs the modest increase in parameter count.

\begin{table}[h!]
    \centering
    \caption{
    \textcolor{black}{Ablation study on the Emotion Projector design comparing our multi-network approach with a single conditional network.}
    }
    \label{tab:ablation_projector_full}
    \begin{tabular}{l | l | c c c}
        \toprule
        \textbf{Settings} & \textbf{Projector Design} & \textbf{FAD\(\downarrow\)} & \textbf{LSE-D\(\downarrow\)} & \textbf{CSIM\(\uparrow\)} \\
        \hline
        \multirow{2}{*}{Inter-ID} & Single Conditional Network & 1.152 & 9.411 & 0.882 \\
        & Multi-Network (Ours) & \textbf{1.075} & \textbf{9.364} & \textbf{0.915} \\
        \hline
        \multirow{2}{*}{Cross-ID} & Single Conditional Network & 4.415 & 9.402 & 0.775 \\
        & Multi-Network (Ours) & \textbf{4.300} & \textbf{9.374} & \textbf{0.784} \\
        \bottomrule
    \end{tabular}
\end{table}

\section{Negative Sampling Strategy Details}
\label{section:negative_sampling}

In our main paper, we introduce a structured negative sampling strategy for the contrastive pre-training of our PEPL module. This section provides the details of this data-driven approach, designed to address the challenge that emotions are not always mutually exclusive (e.g., "anger" and "disgust"), where naively treating any other emotion as a negative sample can lead to a suboptimal learning signal.

\subsection{Empirical Cross-Modal Similarity Analysis}
To inform our sampling strategy, we directly measured the average similarity between the visual representations of each emotion and the standard textual representations of every other emotion. For each of the seven emotion categories, we randomly sampled a large set of images (N=1000) from the MEAD training set and extracted their 512-D visual embeddings using the CLIP-ViT-B/32 image encoder. We then computed the average cosine similarity between these visual embeddings and the single text embedding of each of the seven emotion prompts.

The resulting 7x7 cross-modal similarity matrix is presented in Table~\ref{tab:cross_modal_similarity}. The diagonal entries (bolded) consistently show the highest similarity for each row, confirming that the VLM can generally associate the correct visual and textual concepts.

\begin{table}[h!]
    \centering
    \caption{
    \textcolor{black}{Cross-modal (Image-to-Text) cosine similarity matrix. Each cell (i, j) shows the average similarity between 1,000 images of emotion i and the text prompt for emotion j.}
    }
    \label{tab:cross_modal_similarity}
    \resizebox{\columnwidth}{!}{
    \begin{tabular}{l | c c c c c c c}
        \toprule
        \textbf{Image\textbackslash{}Text} & \textbf{Angry} & \textbf{Disgusted} & \textbf{Fear} & \textbf{Happy} & \textbf{Neutral} & \textbf{Sad} & \textbf{Surprised} \\
        \hline
        \textbf{Angry}     & \textbf{0.452} & 0.418 & 0.315 & 0.221 & 0.289 & 0.334 & 0.321 \\
        \textbf{Disgusted} & 0.409 & \textbf{0.431} & 0.320 & 0.215 & 0.291 & 0.340 & 0.318 \\
        \textbf{Fear}      & 0.311 & 0.318 & \textbf{0.485} & 0.255 & 0.301 & 0.312 & 0.402 \\
        \textbf{Happy}     & 0.225 & 0.218 & 0.258 & \textbf{0.553} & 0.352 & 0.267 & 0.288 \\
        \textbf{Neutral}   & 0.292 & 0.299 & 0.305 & 0.349 & \textbf{0.601} & 0.330 & 0.308 \\
        \textbf{Sad}       & 0.338 & 0.342 & 0.315 & 0.271 & 0.333 & \textbf{0.524} & 0.319 \\
        \textbf{Surprised} & 0.325 & 0.321 & 0.411 & 0.285 & 0.310 & 0.315 & \textbf{0.538} \\
        \bottomrule
    \end{tabular}
    }
\end{table}

\subsection{Pre-defined Negative Sampling Pools}
The empirical analysis in Table~\ref{tab:cross_modal_similarity} provides a direct, data-driven basis for our negative sampling strategy. The table reveals that for each image emotion category (each row), there are one or two non-corresponding text prompts that exhibit significantly higher similarity than the others, representing the most confusable negative samples. To ensure a clear and stable learning signal, our structured strategy is to exclude the top-K most similar non-corresponding text prompts from the negative sampling pool for each anchor image class. 

Based on empirical validation, we found that setting K=1 (i.e., excluding only the single most confusable negative) provided the best balance between removing ambiguity and maintaining a diverse set of hard negatives. The specific pools derived from this Top-1 exclusion strategy and used in all our experiments are detailed in Table~\ref{tab:negative_pools}.

\begin{table}[h!]
    \centering
    \caption{
    \textcolor{black}{Pre-defined negative sampling pools for each positive emotion.}
    }
    \label{tab:negative_pools}
    \begin{tabular}{l | p{6cm}}
        \toprule
        \textbf{Positive Emotion} & \textbf{Valid Negative Sampling Pool} \\
        \hline
        Angry     & \{Fear, Happy, Neutral, Sad, Surprised\} \\
        Disgusted & \{Fear, Happy, Neutral, Sad, Surprised\} \\
        Fear      & \{Angry, Disgusted, Happy, Neutral, Sad\} \\
        Happy     & \{Angry, Disgusted, Fear, Sad, Surprised\} \\
        Neutral   & \{Angry, Disgusted, Fear, Happy, Sad, Surprised\} \\
        Sad       & \{Angry, Fear, Happy, Neutral, Surprised\} \\
        Surprised & \{Angry, Disgusted, Fear, Happy, Neutral, Sad\} \\
        \bottomrule
    \end{tabular}
\end{table}

\section{Analysis of Inconsistent and Marginal Results}
\label{section:inconsistent_results}

In our main paper, we present a comprehensive evaluation of our PCMECL framework. While the results demonstrate broad effectiveness, this section provides a more in-depth, critical analysis of the few cases where our method yields marginal or inconsistent improvements. 

These cases include, for instance, the slight FAD degradation for the "Neutral" category in some settings, and the consistently more marginal gains on the ICface baseline compared to NED. These outcomes are attributed to the complex interaction of the following three key factors:

\begin{itemize}
    \item \textbf{Metric Sensitivity on Saturated Tasks.} For low-variance categories like "Neutral," where baselines already perform exceptionally well, the FAD metric is highly sensitive to any minor, even if perceptually reasonable, variance introduced by our method. For instance, in Table I of our main paper (Cross-ID), the FAD for NED slightly increases from 2.022 to 2.054. We argue this is an artifact of the metric on a saturated task, not a semantic failure.
    
    \item \textbf{Inherent Ambiguity of Emotions.} Subtle emotions like "Fear" and "Disgust" have a more diverse and less distinct visual manifestation. This inherent ambiguity can limit the precision of the VLM's supervisory signal itself. This is reflected not only in less pronounced CSIM gains compared to iconic emotions like "Happy," but it also indirectly impacts FAD and LSE-D, as a less precise emotional target provides a weaker constraint against minor generation artifacts.
    
    \item \textbf{Representational Bottlenecks of Baselines.} The ultimate performance is constrained by the baseline model's capacity. The consistently more marginal gains on ICface compared to NED exemplify this. NED operates on a structured 3DMM parameter space where our signal can optimize effectively. In contrast, ICface, as a single-stage GAN operating in a highly entangled latent space and governed by a sensitive adversarial equilibrium, presents a greater challenge for an external supervisory signal to provide guidance without subtle disruptions.
\end{itemize}

\begin{table*}[!t]
\vspace{-2pt}
\caption{
\textcolor{black}{
Detailed comparison results of FAD, CSIM, and LSE-D on the MEAD dataset. We compare NED, ICface, and SSERD with and without PCMECL under Inter-ID and Cross-ID settings.
}
}
\label{table:supp_mead_detailed}
\centering
\scriptsize
\renewcommand{\arraystretch}{1.2}
\setlength{\tabcolsep}{8pt}
\begin{tabular}{c | c | l || c c c c c c c | c}
\toprule
\textbf{Settings} & \textbf{Metrics} & \textbf{Methods} & \textbf{Neutral} & \textbf{Angry} & \textbf{Disgusted} & \textbf{Fear} & \textbf{Happy} & \textbf{Sad} & \textbf{Surprised} & \textbf{Avg.} \\
\hline \hline
\multirow{18}{*}{\textbf{Inter-ID}} & \multirow{6}{*}{FAD\(\downarrow\)} 
 & NED & 0.906 & 2.177 & 3.838 & 1.659 & 1.939 & 2.538 & 1.700 & 2.108 \\
 & & \textbf{Ours(NED)} & 0.966 & 1.163 & 1.123 & 1.378 & 1.034 & 0.880 & 0.978 & \textbf{1.075} \\
 \cline{3-11}
 & & ICface & 7.114 & 6.420 & 7.383 & 6.567 & 6.213 & 7.301 & 6.567 & 6.795 \\
 & & \textbf{Ours(ICface)} & 6.911 & 5.709 & 7.073 & 6.562 & 6.012 & 6.892 & 7.070 & \textbf{6.604} \\
 \cline{3-11}
 & & SSERD & 0.473 & 0.683 & 0.820 & 0.781 & 0.648 & 0.808 & 0.965 & 0.740 \\
 & & \textbf{Ours(SSERD)} & 0.485 & 0.503 & 0.710 & 0.811 & 0.510 & 0.700 & 0.866 & \textbf{0.655} \\
\cline{2-11}
 & \multirow{6}{*}{LSE-D\(\downarrow\)} 
 & NED & 9.264 & 9.579 & 9.128 & 10.172 & 9.137 & 9.074 & 9.821 & 9.454 \\
 & & \textbf{Ours(NED)} & 9.338 & 9.627 & 9.138 & 9.502 & 9.313 & 9.467 & 9.163 & \textbf{9.364} \\
 \cline{3-11}
 & & ICface & 9.760 & 10.483 & 10.433 & 9.855 & 10.180 & 10.017 & 9.851 & 10.083 \\
 & & \textbf{Ours(ICface)} & 9.111 & 10.086 & 9.908 & 9.604 & 9.219 & 9.437 & 9.598 & \textbf{9.566} \\
 \cline{3-11}
 & & SSERD & 8.964 & 9.188 & 8.796 & 9.181 & 9.315 & 9.253 & 9.188 & 9.126 \\
 & & \textbf{Ours(SSERD)} & 8.850 & 9.043 & 8.853 & 9.050 & 9.204 & 9.150 & 9.116 & \textbf{9.038} \\
\cline{2-11}
 & \multirow{6}{*}{CSIM\(\uparrow\)} 
 & NED & 0.883 & 0.802 & 0.772 & 0.848 & 0.839 & 0.812 & 0.864 & 0.831 \\
 & & \textbf{Ours(NED)} & 0.907 & 0.895 & 0.931 & 0.906 & 0.927 & 0.926 & 0.916 & \textbf{0.915} \\
 \cline{3-11}
 & & ICface & 0.779 & 0.741 & 0.805 & 0.754 & 0.775 & 0.755 & 0.817 & 0.775 \\
 & & \textbf{Ours(ICface)} & 0.801 & 0.776 & 0.788 & 0.794 & 0.834 & 0.765 & 0.799 & \textbf{0.794} \\
 \cline{3-11}
 & & SSERD & 0.908 & 0.874 & 0.929 & 0.892 & 0.908 & 0.914 & 0.906 & 0.904 \\
 & & \textbf{Ours(SSERD)} & 0.909 & 0.885 & 0.925 & 0.870 & 0.940 & 0.925 & 0.902 & \textbf{0.908} \\
\hline
\hline
\multirow{18}{*}{\textbf{Cross-ID}} & \multirow{6}{*}{FAD\(\downarrow\)} 
 & NED & 2.022 & 4.851 & 5.094 & 4.983 & 3.919 & 5.665 & 4.600 & 4.448 \\
 & & \textbf{Ours(NED)} & 2.054 & 4.715 & 4.643 & 4.818 & 3.820 & 5.422 & 4.626 & \textbf{4.300} \\
 \cline{3-11}
 & & ICface & 10.560 & 9.470 & 9.230 & 9.122 & 8.493 & 10.364 & 9.541 & 9.540 \\
 & & \textbf{Ours(ICface)} & 9.568 & 9.485 & 10.134 & 9.134 & 8.613 & 9.465 & 9.291 & \textbf{9.384} \\
 \cline{3-11}
 & & SSERD & 0.490 & 3.308 & 4.011 & 2.215 & 2.307 & 2.483 & 2.360 & 2.453 \\
 & & \textbf{Ours(SSERD)} & 0.604 & 3.650 & 3.247 & 2.200 & 2.799 & 2.300 & 2.287 & \textbf{2.441} \\
\cline{2-11}
& \multirow{6}{*}{LSE-D\(\downarrow\)} 
 & NED & 9.812 & 9.904 & 10.121 & 9.741 & 9.936 & 10.179 & 9.646 & 9.906 \\
 & & \textbf{Ours(NED)} & 9.422 & 9.193 & 9.352 & 9.239 & 9.332 & 9.827 & 9.250 & \textbf{9.374} \\
 \cline{3-11}
 & & ICface & 11.226 & 11.073 & 11.184 & 11.204 & 11.322 & 11.526 & 11.133 & 11.238 \\
 & & \textbf{Ours(ICface)} & 10.699 & 10.034 & 10.192 & 10.260 & 10.078 & 10.304 & 10.260 & \textbf{10.261} \\
 \cline{3-11}
 & & SSERD & 8.956 & 9.016 & 9.051 & 9.364 & 9.304 & 9.299 & 9.410 & 9.200 \\
 & & \textbf{Ours(SSERD)} & 8.850 & 9.131 & 9.066 & 9.250 & 9.150 & 9.083 & 9.121 & \textbf{9.093} \\
\cline{2-11}
 & \multirow{6}{*}{CSIM\(\uparrow\)} 
 & NED & 0.841 & 0.717 & 0.791 & 0.750 & 0.842 & 0.691 & 0.780 & 0.773 \\
 & & \textbf{Ours(NED)} & 0.824 & 0.729 & 0.811 & 0.759 & 0.838 & 0.750 & 0.780 & \textbf{0.784} \\
 \cline{3-11}
 & & ICface & 0.705 & 0.648 & 0.637 & 0.727 & 0.717 & 0.664 & 0.721 & 0.688 \\
 & & \textbf{Ours(ICface)} & 0.689 & 0.658 & 0.739 & 0.707 & 0.769 & 0.654 & 0.764 & \textbf{0.711} \\
 \cline{3-11}
 & & SSERD & 0.908 & 0.799 & 0.843 & 0.837 & 0.858 & 0.838 & 0.852 & 0.848 \\
 & & \textbf{Ours(SSERD)} & 0.885 & 0.830 & 0.832 & 0.858 & 0.879 & 0.852 & 0.863 & \textbf{0.857} \\
\bottomrule
\end{tabular}
\end{table*}

\begin{table*}[!t]
\vspace{-2pt}
\caption{
\textcolor{black}{
Detailed comparison results of FAD, CSIM, and LSE-D on the RAVDESS dataset. We compare NED, ICface, and SSERD with and without PCMECL under Inter-ID and Cross-ID settings.
}
}
\label{table:supp_ravdess_detailed}
\centering
\scriptsize
\renewcommand{\arraystretch}{1.2}
\setlength{\tabcolsep}{8pt}
\begin{tabular}{c | c | l || c c c c c c c | c}
\toprule
\textbf{Settings} & \textbf{Metrics} & \textbf{Methods} & \textbf{Neutral} & \textbf{Angry} & \textbf{Disgusted} & \textbf{Fear} & \textbf{Happy} & \textbf{Sad} & \textbf{Surprised} & \textbf{Avg.} \\
\hline \hline
\multirow{18}{*}{\textbf{Inter-ID}} & \multirow{6}{*}{FAD\(\downarrow\)} 
 & NED & 2.041 & 3.288 & 4.144 & 2.635 & 3.714 & 2.595 & 2.980 & 3.057 \\
 & & \textbf{Ours(NED)} & 2.295 & 3.029 & 3.959 & 1.906 & 3.148 & 2.215 & 2.601 & \textbf{2.736} \\
 \cline{3-11}
 & & ICface & 9.816 & 7.047 & 8.689 & 8.413 & 8.413 & 8.086 & 8.636 & 8.443 \\
 & & \textbf{Ours(ICface)} & 7.340 & 7.883 & 8.095 & 8.349 & 8.738 & 7.679 & 7.717 & \textbf{7.972} \\
 \cline{3-11}
 & & SSERD & 1.021 & 1.127 & 1.287 & 1.483 & 1.161 & 1.999 & 1.718 & 1.399 \\
 & & \textbf{Ours(SSERD)} & 1.102 & 1.050 & 1.301 & 1.220 & 1.189 & 1.648 & 1.562 & \textbf{1.296} \\
\cline{2-11}
 & \multirow{6}{*}{LSE-D\(\downarrow\)} 
 & NED & 7.376 & 7.757 & 7.822 & 7.452 & 7.742 & 7.560 & 7.226 & 7.562 \\
 & & \textbf{Ours(NED)} & 7.372 & 7.603 & 7.216 & 7.316 & 7.724 & 7.432 & 7.558 & \textbf{7.460} \\
 \cline{3-11}
 & & ICface & 8.209 & 9.504 & 8.295 & 8.523 & 8.902 & 8.346 & 7.578 & 8.480 \\
 & & \textbf{Ours(ICface)} & 7.444 & 9.949 & 9.173 & 8.831 & 7.736 & 7.332 & 7.537 & \textbf{8.286} \\
 \cline{3-11}
 & & SSERD & 7.578 & 7.786 & 7.428 & 7.136 & 7.472 & 7.363 & 7.326 & 7.441 \\
 & & \textbf{Ours(SSERD)} & 7.485 & 7.702 & 7.315 & 7.052 & 7.385 & 7.280 & 7.266 & \textbf{7.355} \\
\cline{2-11}
 & \multirow{6}{*}{CSIM\(\uparrow\)} 
 & NED & 0.847 & 0.805 & 0.786 & 0.842 & 0.793 & 0.855 & 0.848 & 0.825 \\
 & & \textbf{Ours(NED)} & 0.797 & 0.810 & 0.796 & 0.862 & 0.859 & 0.847 & 0.860 & \textbf{0.833} \\
 \cline{3-11}
 & & ICface & 0.749 & 0.703 & 0.775 & 0.722 & 0.797 & 0.766 & 0.772 & 0.755 \\
 & & \textbf{Ours(ICface)} & 0.789 & 0.711 & 0.767 & 0.768 & 0.805 & 0.807 & 0.799 & \textbf{0.778} \\
 \cline{3-11}
 & & SSERD & 0.898 & 0.810 & 0.927 & 0.886 & 0.914 & 0.932 & 0.894 & 0.894 \\
 & & \textbf{Ours(SSERD)} & 0.908 & 0.807 & 0.913 & 0.911 & 0.936 & 0.933 & 0.885 & \textbf{0.899} \\
\hline
\hline
\multirow{18}{*}{\textbf{Cross-ID}} & \multirow{6}{*}{FAD\(\downarrow\)} 
 & NED & 3.558 & 5.546 & 7.388 & 5.008 & 5.648 & 5.588 & 5.145 & 5.412 \\
 & & \textbf{Ours(NED)} & 3.129 & 5.415 & 7.146 & 4.942 & 5.157 & 5.684 & 4.952 & \textbf{5.204} \\
 \cline{3-11}
 & & ICface & 10.478 & 8.704 & 9.260 & 9.106 & 9.061 & 9.639 & 9.718 & 9.424 \\
 & & \textbf{Ours(ICface)} & 10.646 & 8.064 & 8.481 & 9.362 & 8.648 & 9.905 & 9.344 & \textbf{9.207} \\
 \cline{3-11}
 & & SSERD & 1.085 & 4.045 & 5.785 & 3.431 & 3.336 & 3.063 & 2.772 & 3.360 \\
 & & \textbf{Ours(SSERD)} & 1.150 & 4.052 & 5.600 & 3.397 & 3.301 & 3.050 & 2.970 & \textbf{3.360} \\
\cline{2-11}
& \multirow{6}{*}{LSE-D\(\downarrow\)} 
 & NED & 7.856 & 8.085 & 8.107 & 8.151 & 8.073 & 8.006 & 7.962 & 8.034 \\
 & & \textbf{Ours(NED)} & 8.122 & 7.712 & 8.158 & 8.146 & 7.536 & 7.804 & 8.056 & \textbf{7.933} \\
 \cline{3-11}
 & & ICface & 10.736 & 12.415 & 11.860 & 11.279 & 11.150 & 11.305 & 12.028 & 11.539 \\
 & & \textbf{Ours(ICface)} & 9.560 & 11.392 & 10.857 & 10.262 & 10.811 & 10.561 & 10.634 & \textbf{10.582} \\
 \cline{3-11}
 & & SSERD & 7.557 & 7.587 & 7.595 & 7.690 & 7.524 & 7.693 & 7.704 & 7.621 \\
 & & \textbf{Ours(SSERD)} & 7.466 & 7.517 & 7.520 & 7.600 & 7.550 & 7.577 & 7.403 & \textbf{7.519} \\
\cline{2-11}
 & \multirow{6}{*}{CSIM\(\uparrow\)} 
 & NED & 0.820 & 0.766 & 0.741 & 0.749 & 0.804 & 0.726 & 0.713 & 0.760 \\
 & & \textbf{Ours(NED)} & 0.838 & 0.752 & 0.756 & 0.782 & 0.827 & 0.727 & 0.725 & \textbf{0.772} \\
 \cline{3-11}
 & & ICface & 0.677 & 0.646 & 0.717 & 0.649 & 0.738 & 0.666 & 0.644 & 0.677 \\
 & & \textbf{Ours(ICface)} & 0.696 & 0.633 & 0.701 & 0.686 & 0.704 & 0.723 & 0.627 & \textbf{0.681} \\
 \cline{3-11}
 & & SSERD & 0.898 & 0.737 & 0.676 & 0.759 & 0.809 & 0.856 & 0.794 & 0.790 \\
 & & \textbf{Ours(SSERD)} & 0.885 & 0.755 & 0.700 & 0.775 & 0.825 & 0.860 & 0.807 & \textbf{0.801} \\
\bottomrule
\end{tabular}
\end{table*}

\section{Justification for Optimizer and Learning Rate Scheduler Choice}
\label{section:optimizer_choice}
In our main paper, we specify the use of a Stochastic Gradient Descent (SGD) optimizer with a manually-tuned step decay learning rate schedule for the pre-training of our PEPL module. This section provides the justification for this engineering choice over more common adaptive optimizers like Adam.

Our decision was based on empirical findings during initial experiments. We observed that the pre-training of the PEPL module, which involves a contrastive learning objective, presents a particularly challenging optimization problem for two primary reasons:

\begin{itemize}
    \item \textbf{Volatile Loss Landscape:} The loss is computed based on dynamically constructed positive and negative pairs within each mini-batch. This causes sharp fluctuations in the gradient's direction and magnitude between iterations, creating a non-smooth loss landscape.
    
    \item \textbf{Multi-Component Co-optimization:} The PEPL module requires the simultaneous optimization of multiple distinct networks (the Visual Guider's MLP and the seven Emotion Projector MLPs), which have different architectures and potentially different convergence dynamics.
\end{itemize}

In this challenging context, we found that adaptive optimizers like Adam often struggled with instability. Their adaptive momentum mechanism, when faced with a volatile loss landscape, can accumulate excessive update steps, leading to overshooting and training divergence. Furthermore, their single base learning rate can be difficult to tune to simultaneously suit all functionally different components.

In contrast, SGD with momentum, combined with our manual step decay schedule, demonstrated superior robustness and stability in our experiments. The simpler update mechanism of SGD is less susceptible to being misled by transient gradient fluctuations, allowing it to converge reliably. The manual step decay then provided effective macro-level control over the training process.

Therefore, our choice was an empirically-driven decision to prioritize training stability and final model performance for our specific, challenging task.

\section{\textcolor{black}{Detailed Quantitative Analysis}}
\label{sec:detailed_quant}
\textcolor{black}{
To provide a more comprehensive evaluation, this section presents the detailed breakdown of performance metrics (FAD, LSE-D, and CSIM) for each specific emotion (e.g., Neutral, Angry, Disgusted, Fear, Happy, Sad, Surprised) on both the MEAD and RAVDESS datasets.
}

\textcolor{black}{
We evaluate the effectiveness of our proposed PCMECL framework by integrating it into the representative baselines (NED, ICface) and the state-of-the-art method SSERD. We compare the performance with and without our supervision under both Inter-ID and Cross-ID settings. As illustrated in Table \ref{table:supp_mead_detailed} and Table \ref{table:supp_ravdess_detailed}, our method demonstrates superior performance. Notably, even for challenging emotions with subtle visual cues or large deformations, our PCMECL-supervised models generally outperform their respective baselines, particularly in emotional accuracy (CSIM). This confirms that the improvements reported in the main text are robust and broadly applicable across diverse emotional expressions.
}

\begin{figure}[t]
\centering
\includegraphics[width=\columnwidth]{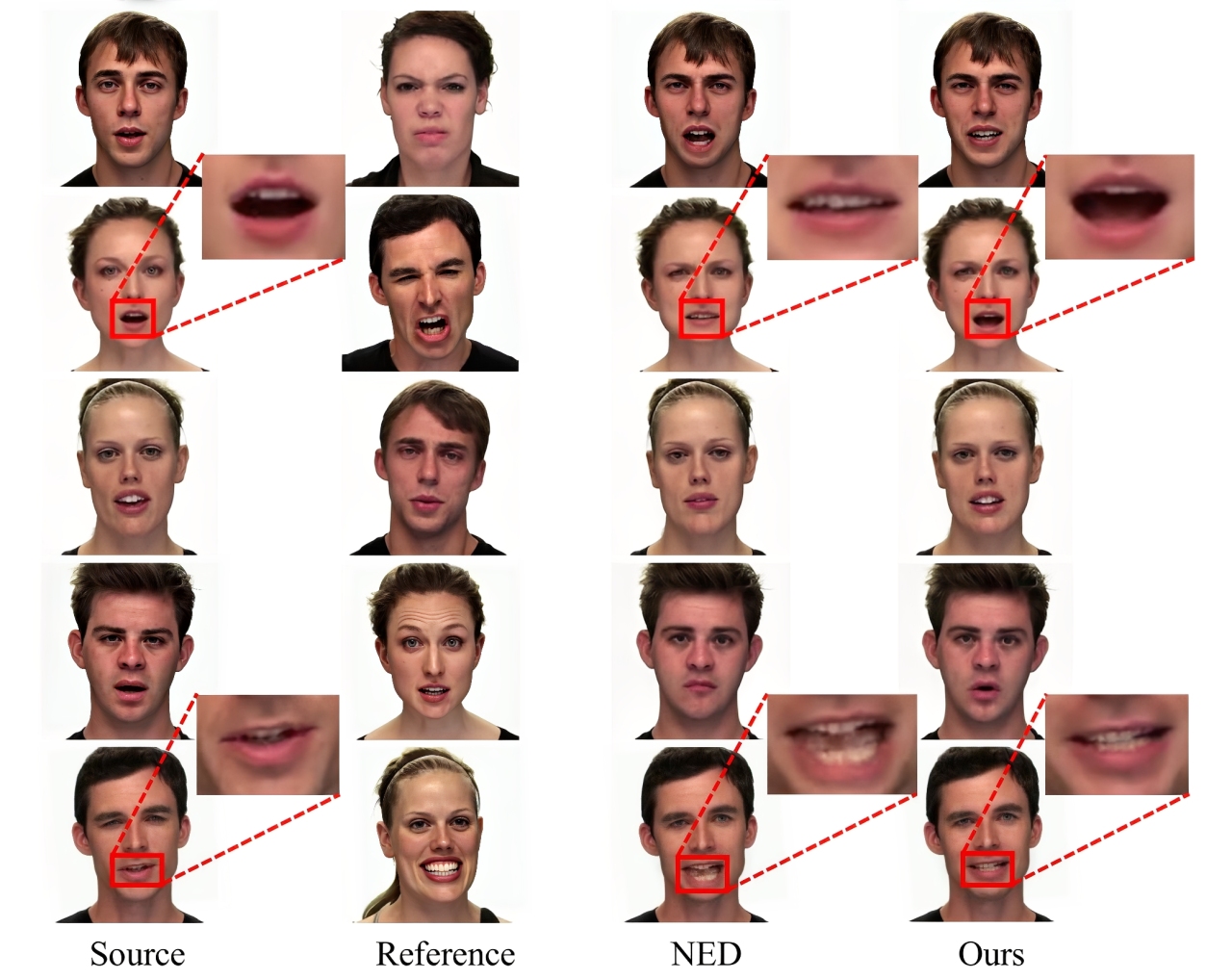} 
\caption{
\textcolor{black}{
Qualitative comparisons of NED with and without PCMECL supervision on the RAVDESS dataset.
}
}
\label{fig: NED_RAVDESS_comparison}
\end{figure}

\begin{figure}[t]
\centering
\includegraphics[width=\columnwidth]{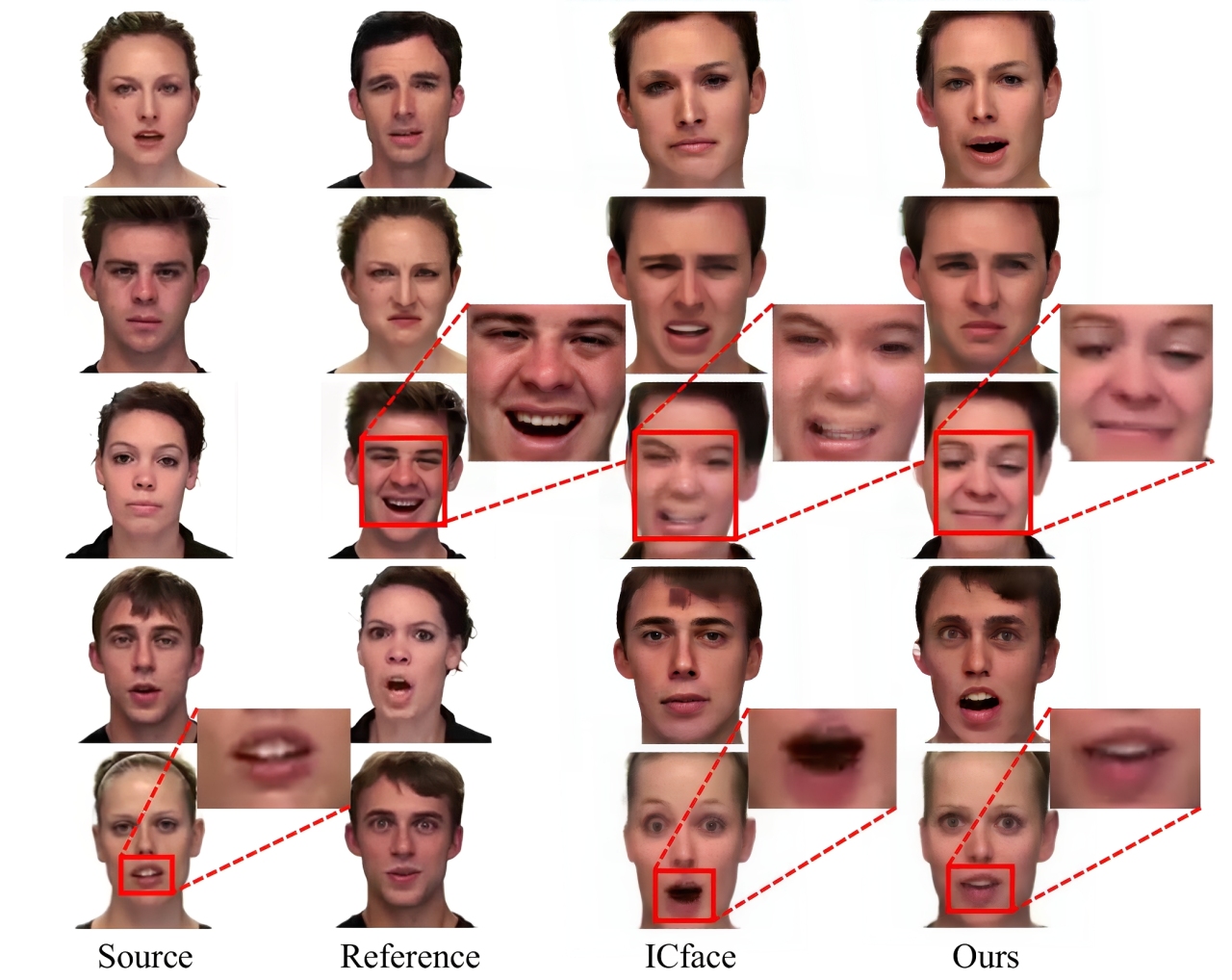} 
\caption{
\textcolor{black}{
Qualitative comparisons of ICface with and without PCMECL supervision on the RAVDESS dataset.
}
}
\label{fig: ICface_RAVDESS_comparison}
\end{figure}

\begin{figure}[t]
\centering
\includegraphics[width=\columnwidth]{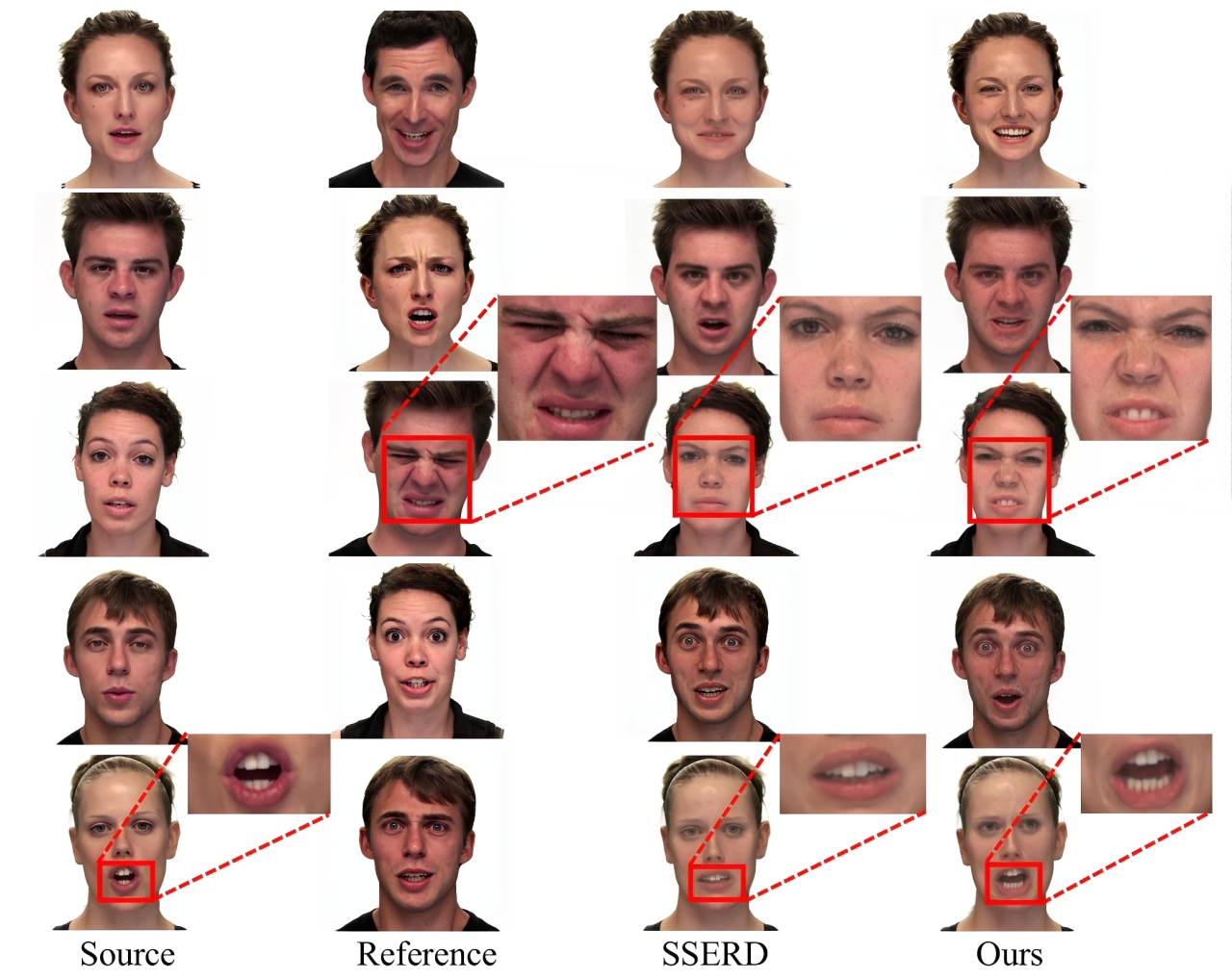} 
\caption{
\textcolor{black}{
Qualitative comparisons of SSERD with and without PCMECL supervision on the RAVDESS dataset.
}
}
\label{fig: SSERD_RAVDESS_comparison}
\end{figure}

\section{\textcolor{black}{Qualitative Comparisons on the RAVDESS Dataset}}
\label{sec:supp_qualitative}
\textcolor{black}{To further evaluate the generalization capability, we provide additional qualitative results on the RAVDESS dataset. We compare our method against the representative baselines (NED, ICface) and the state-of-the-art method (SSERD) in Fig. \ref{fig: NED_RAVDESS_comparison}, Fig. \ref{fig: ICface_RAVDESS_comparison}, and Fig. \ref{fig: SSERD_RAVDESS_comparison}, respectively. In these visualizations, we transfer the emotion from a Reference image to a Source identity. The `Ours' column displays the final results supervised by PCMECL, where red boxes and corresponding zoom-in panels are provided to highlight the superior detail in key expressive regions.
}

\textcolor{black}{
As shown in Fig. \ref{fig: NED_RAVDESS_comparison} and Fig. \ref{fig: ICface_RAVDESS_comparison}, the representative baselines (NED and ICface) often producing blurred faces or significant artifacts. In contrast, the SOTA method SSERD (Fig. \ref{fig: SSERD_RAVDESS_comparison}) maintains better structural integrity but occasionally suffers from semantic drift, failing to capture the specific nuance of the target emotion. Our method consistently generates high-fidelity expressions while robustly preserving the source identity, demonstrating superior zero-shot generalization.
}

\clearpage

\bibliographystyle{IEEEtran}
\bibliography{reference}

\begin{IEEEbiography}[{\includegraphics[width=1in,height=1.25in,clip,keepaspectratio]{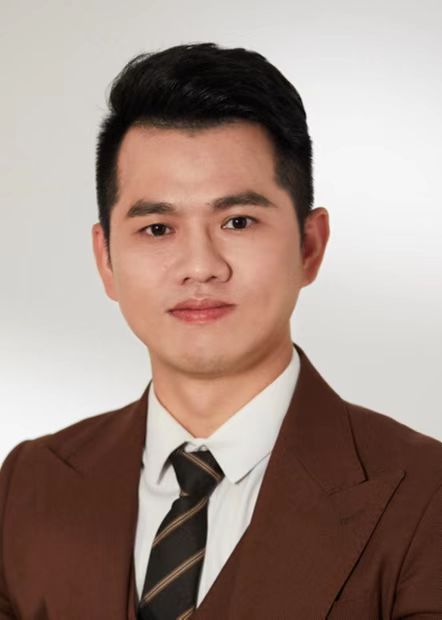}}]{Tianshui Chen} received a Ph.D. degree in computer science at the School of Data and Computer Science Sun Yat-sen University, Guangzhou, China, in 2018. Prior to earning his Ph.D, he received a B.E. degree from the School of Information and Science Technology in 2013. He is currently an associate professor at the Guangdong University of Technology. His current research interests include artificial intelligence, multimodal large models, and generative AI. He has authored and co-authored more than 60 papers published in top-tier academic journals and conferences, including T-PAMI, IJCV, T-NNLS, T-IP, T-MM, CVPR, ICCV, AAAI, IJCAI, ACM MM, etc. He has served as a reviewer for numerous academic journals and conferences. He was the recipient of the Best Paper Diamond Award at IEEE ICME 2017. 
\end{IEEEbiography}

\begin{IEEEbiography}[{\includegraphics[width=1in,height=1.25in,clip,keepaspectratio]{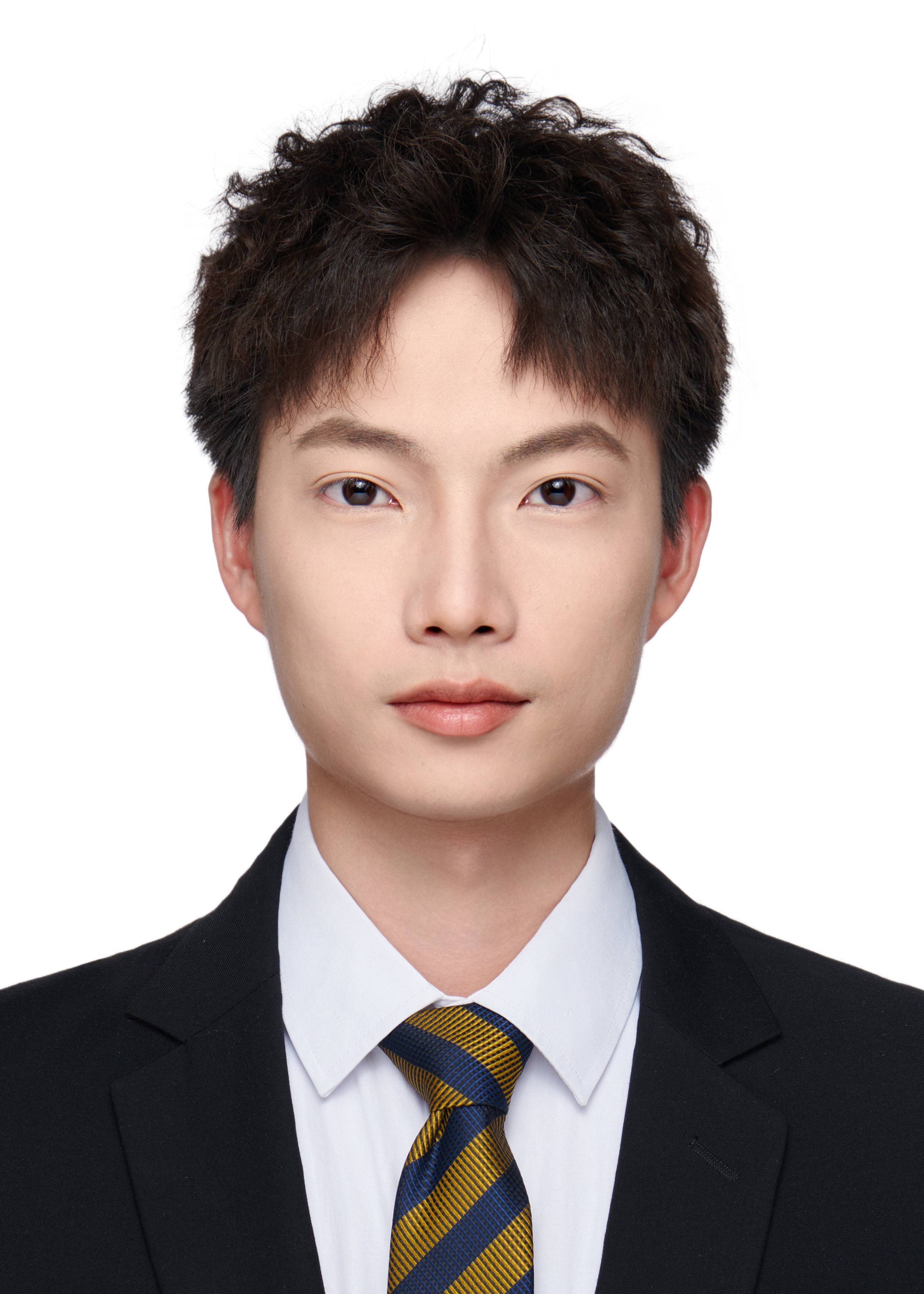}}]{Yujie Zhu} received the B.Sc. degree in Communications Engineering from Dongguan University of Technology, Dongguan, China, in 2023. He is currently pursuing the M.Eng. degree in Electronic Information at Guangdong University of Technology, Guangzhou, China. His research focuses on artificial intelligence, multimodal large-scale models, and generative AI. 
\end{IEEEbiography}

\begin{IEEEbiography}[{\includegraphics[width=1in,height=1.25in,clip,keepaspectratio]{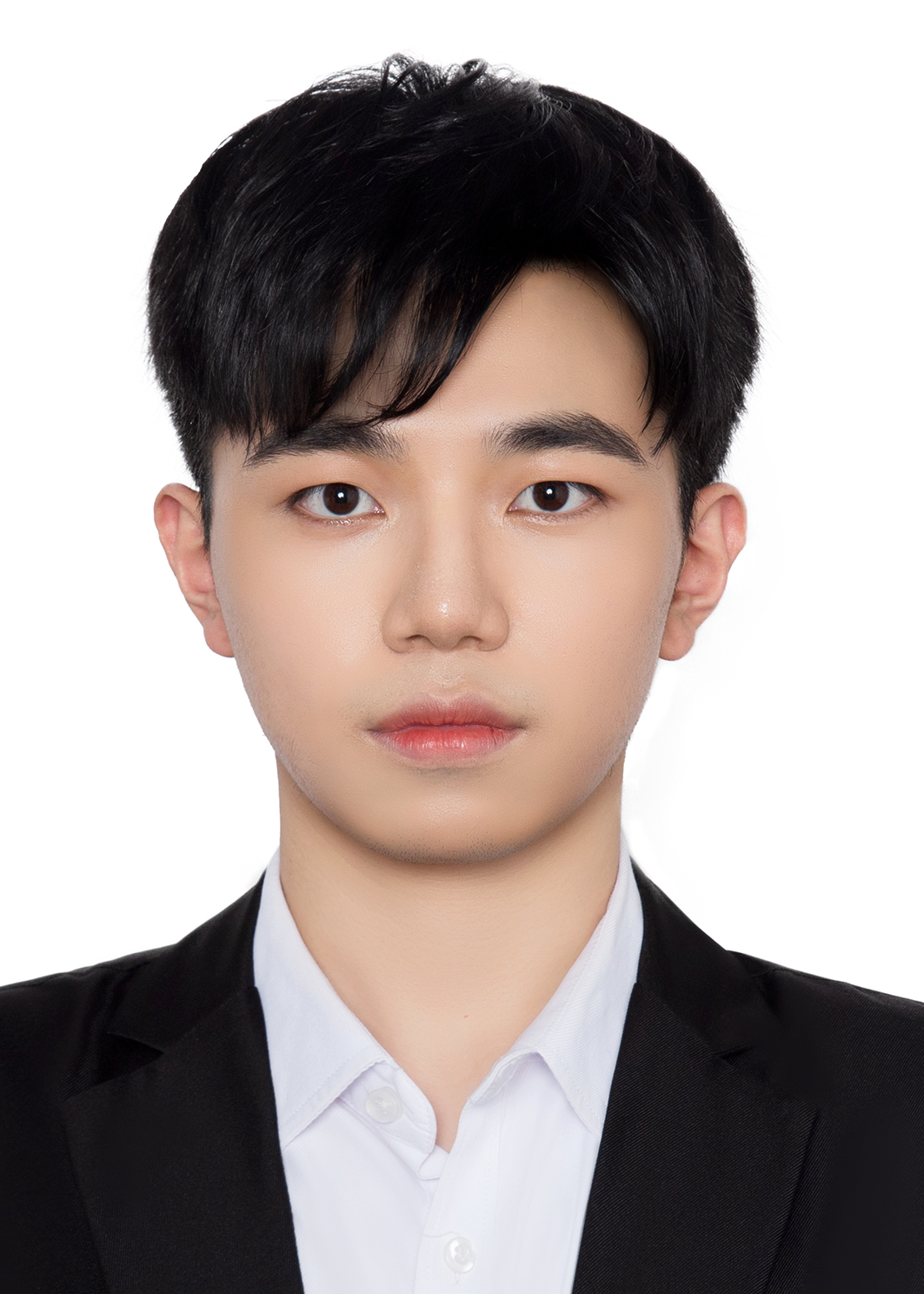}}]{Jianman Lin} received a B.Sc. degree from Guangdong University of Technology, Guangzhou, China, in 2024. He is currently pursuing a Master's degree in Electronic Information at South China University of Technology. His research interests include artificial intelligence, multimodal large models, and generative AI. He has published revised papers in prestigious conferences and journals, including CVPR and IJCV.
\end{IEEEbiography}

\begin{IEEEbiography}[{\includegraphics[width=1in,height=1.25in,clip,keepaspectratio]{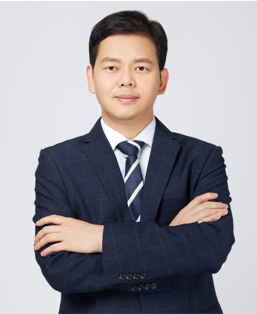}}]{Zhijing Yang} received the B.S and Ph.D. degrees from the Mathematics and Computing Science, Sun Yat-sen University, Guangzhou China, in 2003 and 2008, respectively. He was a Visiting Research Scholar in the School of Computing, Informatics and Media, University of Bradford, U.K, between July-Dec, 2009, and a Research Fellow in the School of Engineering, University of Lincoln, U.K, between Jan. 2011 to Jan. 2013. He is currently a Professor and Vice Dean at the School of Information Engineering, Guangdong University of Technology, China. He has published over 80 peer-reviewed journal and conference papers, including IEEE T-CSVT, T-MM, T-GRS, PR, etc. His research interests include machine learning and pattern recognition.
\end{IEEEbiography}

\begin{IEEEbiography}[{\includegraphics[width=1in,height=1.25in,clip,keepaspectratio]{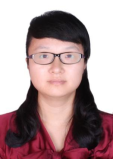}}]{Chunmei Qing}  received her B.Sc. degree in Information and Computation Science from Sun Yat-sen University, China, in 2003, and Ph.D. degree in Electronic Imaging and Media Communications from University of Bradford, UK, in 2009. Then she worked as a postdoctoral researcher in the University of Lincoln, UK. From 2013 till now, she is an associate professor in School of Electronic and Information Engineering, South China University of Technology (SCUT), Guangzhou, China. Her main research interests include image/video processing, computer vision, and affective computing.
\end{IEEEbiography}

\begin{IEEEbiography}[{\includegraphics[width=1in,clip]{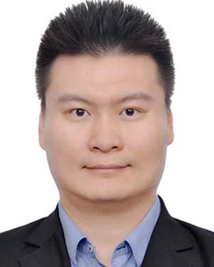}}]{Gao Feng} received the BS degree in computer science from University College London, in 2007, and the PhD degree in computer science from Peking University, in 2018. He was a post-doctoral research fellow with the Future Laboratory, Tsinghua University, from 2018 to 2020. He joins Peking University as Assistant Professor since 2020. His research interest is on the intersection of computer science and art, including but not limit on artificial intelligence and painting art, deep learning and painting robot, etc.
\end{IEEEbiography}

\begin{IEEEbiography}[{\includegraphics[width=1in,clip]{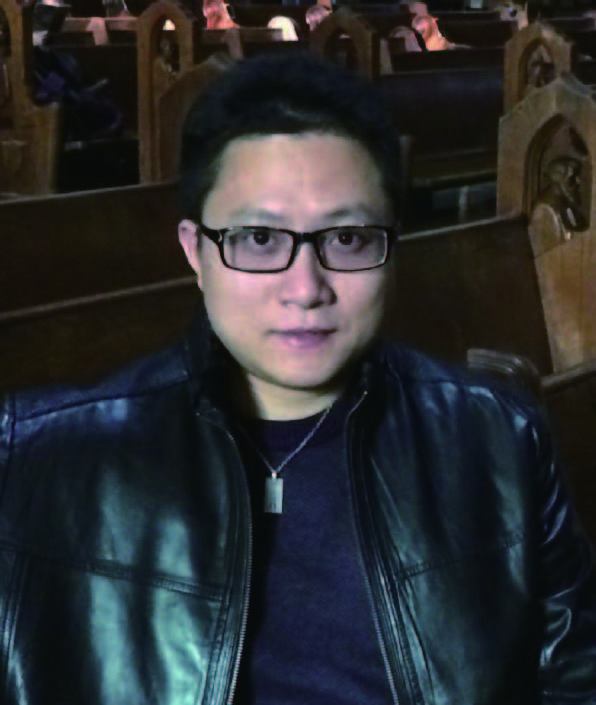}}]{Liang Lin} (Fellow, IEEE) is a full professor at Sun Yat-sen University. From 2008 to 2010, he was a postdoctoral fellow at the University of California, Los Angeles. From 2016--2018, he led the SenseTime R\&D teams to develop cutting-edge and deliverable solutions for computer vision, data analysis and mining, and intelligent robotic systems. He has authored and co-authored more than 100 papers in top-tier academic journals and conferences (e.g., 15 papers in TPAMI and IJCV and 60+ papers in CVPR, ICCV, NIPS, and IJCAI). He has served as an associate editor of IEEE Trans. Human-Machine Systems, The Visual Computer, and Neurocomputing and as an area/session chair for numerous conferences, such as CVPR, ICME, ACCV, and ICMR. He was the recipient of the Annual Best Paper Award by Pattern Recognition (Elsevier) in 2018, the Best Paper Diamond Award at IEEE ICME 2017, the Best Paper Runner-Up Award at ACM NPAR 2010, Google Faculty Award in 2012, the Best Student Paper Award at IEEE ICME 2014, and the Hong Kong Scholars Award in 2014. He is a Fellow of IEEE, IAPR, and IET. 
\end{IEEEbiography}

\vfill

\end{document}